%% file: iclr2023_conference.tex
\documentclass{article} 
\usepackage{iclr2023_conference,times}

\input{math_commands.tex}

\usepackage{hyperref}
\usepackage{url}

\usepackage[utf8]{inputenc} 
\usepackage[T1]{fontenc}    
\usepackage{hyperref}       
\usepackage{url}            
\usepackage{booktabs}       
\usepackage{amsfonts}       
\usepackage{nicefrac}       
\usepackage{microtype}      
\usepackage{xcolor}         

\usepackage{times}
\usepackage{latexsym}
\usepackage{amsmath}
\usepackage{enumitem}

\usepackage{pgfplots}
\usetikzlibrary{matrix}
\usepgfplotslibrary{groupplots}
\pgfplotsset{compat=newest}
\pgfplotsset{compat=1.11,
    /pgfplots/ybar legend/.style={
    /pgfplots/legend image code/.code={%
       \draw[##1,/tikz/.cd,yshift=-0.25em]
        (0cm,0cm) rectangle (3pt,0.8em);},
   },
}

\usepackage{listings}

\newcommand{\scale}{0.91}

\usepackage{tikz}
\usetikzlibrary{tikzmark}
\usepackage{listings}
\usepackage[scaled=0.8]{FiraMono}
\usepackage{xcolor}
\usepackage[compatibility=false]{caption}

\definecolor{lightred}{rgb}{1, 0.7, 0.7}
\definecolor{lightgreen}{rgb}{0.7, 1, 0.7}

\definecolor{codegreen}{rgb}{0,0.6,0}
\definecolor{codegray}{rgb}{0.5,0.5,0.5}
\definecolor{codepurple}{rgb}{0.58,0,0.82}
\definecolor{backcolour}{rgb}{0.95,0.95,0.92}
\lstdefinestyle{mystyle}{
    backgroundcolor=\color{white},   
    commentstyle=\color{codegreen},
    keywordstyle=\color{magenta},
    numberstyle=\tiny\color{codegray},
    stringstyle=\color{codepurple},
    basicstyle=\ttfamily\footnotesize,
    breakatwhitespace=false,         
    breaklines=true,                 
    captionpos=b,                    
    keepspaces=true,                 
    numbersep=5pt,                  
    showspaces=false,                
    showstringspaces=false,
    showtabs=false,                  
    tabsize=2,
    escapechar=|,
}
\makeatletter

  \newcount\bt@rangea
  \newcount\bt@rangeb

  \newcommand\btIfInRange[2]{%
      \global\let\bt@inrange\@secondoftwo%
      \edef\bt@rangelist{#2}%
      \foreach \range in \bt@rangelist {%
          \afterassignment\bt@getrangeb%
          \bt@rangea=0\range\relax%
          \pgfmathtruncatemacro\result{ ( #1 >= \bt@rangea) && (#1 <= \bt@rangeb) }%
          \ifnum\result=1\relax%
              \breakforeach%
              \global\let\bt@inrange\@firstoftwo%
          \fi%
      }%
      \bt@inrange%
  }
  \newcommand\bt@getrangeb{%
      \@ifnextchar\relax%
          {\bt@rangeb=\bt@rangea}%
          {\@getrangeb}%
  }
  \def\@getrangeb-#1\relax{%
      \ifx\relax#1\relax%
          \bt@rangeb=100000
      \else%
          \bt@rangeb=#1\relax%
      \fi%
  }

  \newcommand{\btLstHL}[2]{%
    \btIfInRange{\value{lstnumber}}{#1}{#2}{}
  }%

   \makeatletter
   \let\old@lstKV@SwitchCases\lstKV@SwitchCases
   \def\lstKV@SwitchCases#1#2#3{}
   \makeatother
   \usepackage{lstlinebgrd}
   \makeatletter
   \let\lstKV@SwitchCases\old@lstKV@SwitchCases
   
   \lst@Key{numbers}{none}{%
       \def\lst@PlaceNumber{\lst@linebgrd\ }%
       \lstKV@SwitchCases{#1}%
       {none:\\%
        left:\def\lst@PlaceNumber{\llap{\normalfont
                   \lst@numberstyle{\thelstnumber}\kern\lst@numbersep}\lst@linebgrd}\\%
        right:\def\lst@PlaceNumber{\rlap{\normalfont
                   \kern\linewidth \kern\lst@numbersep
                   \lst@numberstyle{\thelstnumber}}\lst@linebgrd}%
       }{\PackageError{Listings}{Numbers #1 unknown}\@ehc}}
   \makeatother

\definecolor{l1}{HTML}{1b9e77}
\definecolor{l2}{HTML}{d95f02}
\definecolor{l3}{HTML}{7570b3}
\definecolor{ll1}{HTML}{e7fef7}
\definecolor{ll2}{HTML}{feede6}

\usepackage{microtype}
\usepackage{pgfplots}
\usepackage{multirow}
\usepackage{graphicx}
\usepackage{xspace}

\newcommand{\mtpb}{{MTPB}\xspace} 
\newcommand{\mtpbf}{{Multi-Turn Programming Benchmark}\xspace} 
\newcommand{\gptnt}{\textsc{GPT-Neo 350M}\xspace}
\newcommand{\gptns}{\textsc{GPT-Neo 2.7B}\xspace}
\newcommand{\gptn}{\textsc{GPT-Neo}\xspace}
\newcommand{\gptj}{\textsc{GPT-J}\xspace}
\newcommand{\model}{\textsc{CodeGen}\xspace}
\newcommand{\nl}{{\textsc{CodeGen-NL}}\xspace}
\newcommand{\pl}{{\textsc{CodeGen-Multi}}\xspace}
\newcommand{\mpl}{{\textsc{CodeGen-Mono}}\xspace}
\newcommand{\nlt}{{\textsc{CodeGen-NL 350M}}\xspace}
\newcommand{\nls}{{\textsc{CodeGen-NL 2.7B}}\xspace}
\newcommand{\nll}{{\textsc{CodeGen-NL 6.1B}}\xspace}
\newcommand{\plt}{{\textsc{CodeGen-Multi 350M}}\xspace}
\newcommand{\pls}{{\textsc{CodeGen-Multi 2.7B}}\xspace}
\newcommand{\pll}{{\textsc{CodeGen-Multi 6.1B}}\xspace}
\newcommand{\mplt}{{\textsc{CodeGen-Mono 350M}}\xspace}
\newcommand{\mpls}{{\textsc{CodeGen-Mono 2.7B}}\xspace}
\newcommand{\mpll}{{\textsc{CodeGen-Mono 6.1B}}\xspace}
\newcommand{\nlxl}{{\textsc{CodeGen-NL 16.1B}}\xspace}
\newcommand{\plxl}{{\textsc{CodeGen-Multi 16.1B}}\xspace}
\newcommand{\mplxl}{{\textsc{CodeGen-Mono 16.1B}}\xspace}
\newcommand{\codext}{\textsc{Codex 300M}\xspace}
\newcommand{\codexs}{\textsc{Codex 2.5B}\xspace}
\newcommand{\codexl}{\textsc{Codex 12B}\xspace}
\newcommand{\jf}{\textsc{JAXformer}\xspace}
\newcommand{\dnl}{\textsc{ThePile}\xspace}
\newcommand{\dpl}{\textsc{BigQuery}\xspace}
\newcommand{\dmpl}{\textsc{BigPython}\xspace}

\usepackage{soul}
\usepackage{float}

\newcommand\extrafootertext[1]{%
    \bgroup
    \renewcommand\thefootnote{\fnsymbol{footnote}}%
    \renewcommand\thempfootnote{\fnsymbol{mpfootnote}}%
    \footnotetext[0]{#1}%
    \egroup
}

\newcommand\legal[1]{#1}

\usepackage{tikz}
\DeclareRobustCommand\circled[1]{\tikz[baseline=(char.base)]{
            \node[shape=circle,draw,inner sep=1pt, thick=1] (char) {#1};}}

\title{CodeGen: An Open Large Language Model for Code with Multi-Turn Program Synthesis}

\author{Erik Nijkamp\thanks{\hspace{4pt}Equal contribution.},\enspace{\bf Bo Pang}\footnotemark[1],\enspace{\bf Hiroaki Hayashi}\footnotemark[1],\vspace{4pt}\\
{\bf Lifu Tu},\enspace{\bf Huan Wang},\enspace{\bf Yingbo Zhou},\enspace{\bf Silvio Savarese},\enspace{\bf Caiming Xiong}\\
\\Salesforce Research}

\iclrfinalcopy
\begin{document}
\maketitle
\extrafootertext{Correspondence to:
Erik Nijkamp (\href{mailto:erik.nijkamp@salesforce.com}{erik.nijkamp@salesforce.com}), Bo Pang (\href{mailto:b.pang@salesforce.com}{b.pang@salesforce.com}), Hiroaki Hayashi (\href{mailto:hiroakihayashi@salesforce.com }{hiroakihayashi@salesforce.com}), Yingbo Zhou (\href{mailto:yingbo.zhou@salesforce.com}{yingbo.zhou@salesforce.com}), Caiming Xiong (\href{mailto:cxiong@salesforce.com}{cxiong@salesforce.com}).}
\begin{abstract}

Program synthesis strives to generate a computer program as a solution to a given problem specification, expressed with input-output examples or natural language descriptions.
The prevalence of large language models advances the state-of-the-art for program synthesis, though limited training resources and data impede open access to such models.
To democratize this, we train and release a family of large language models up to 16.1B parameters, called \mbox{\model}, on natural language and programming language data, and open source the training library \mbox{\jf}.
We show the utility of the trained model by demonstrating that it is competitive with the previous state-of-the-art on zero-shot Python code generation on HumanEval.
We further investigate the multi-step paradigm for program synthesis, where a single program is factorized into multiple prompts specifying subproblems.
To this end, we construct an open benchmark, Multi-Turn Programming Benchmark (MTPB), consisting of 115 diverse problem sets that are factorized into multi-turn prompts. 
Our analysis on \mbox{\mtpb} shows that the same intent provided to \mbox{\model} in multi-turn fashion significantly improves program synthesis over that provided as a single turn.
We make the training library \mbox{\jf} and model checkpoints available as open source contribution: \url{https://github.com/salesforce/CodeGen}.
\end{abstract}

\section{Introduction}

Creating a program has typically involved a human entering code by hand. The goal of program synthesis is to automate the coding process, and generate a computer program that satisfies the user's specified intent. Some have called it the holy grail of computer science \citep{manna1971toward, gulwani2017program}. Successful program synthesis would not only improve the productivity of experienced programmers but also make programming accessible to a wider audience. 

Two key challenges arise when striving to achieve program synthesis: (1) the intractability of the search space, and (2) the difficulty of properly specifying user intent.
To maintain an expressive search space, one needs a large search space, which poses challenges in efficient search. Previous work \citep{joshi2002denali, panchekha2015automatically, cheung2013optimizing} leverages domain-specific language to restrict the search space; however, this limits the applicability of synthesized programs. 
On the contrary, while being widely applicable, general-purpose programming languages (\textit{e.g.}, C, Python) introduce an even larger search space for possible programs. To navigate through the enormous program space, we formulate the task as language modeling, learning a conditional distribution of the next token given preceding tokens and leverage transformers~\citep{vaswani2017attention} and large-scale self-supervised pre-training. This approach has seen success across modalities~\citep{devlin2018bert,lewis2020bart,dosovitskiy2020image}.
Likewise, prior works have developed pre-trained language models for programming language understanding~\citep{kanade2020learning, feng2020codebert}.

To realize program synthesis successfully, users must employ some means to communicate their intent to the models such as a logical expression (which specifies a logical relation between inputs and outputs of a program), pseudo-code, input-output examples, or a verbalized specifications in natural language. 
On the one hand, a complete formal specification enjoys the exact specifications of user intent but may require domain expertise and effort from users to translate the intent to such a form.
On the other hand, specification merely based on input-output examples is less costly but may under-specify the intent, leading to inaccurate solutions. 
Previous work has benefited from various methods and their combinations as the input to program synthesis models, including pseudo-code~\citep{kulal2019spoc}, a part of a program and its documentation~\citep{chen2021evaluating}, or natural language paragraph with input-output examples~\citep{hendrycks2021measuring}.
However, we argue that a truly user-friendly form of intent is natural language text.

To overcome these challenges,
we propose a multi-turn program synthesis approach, where a user communicates with the synthesis system 
by progressively providing specifications in natural language while receiving responses from the system in the form of synthesized subprograms, such that the user together with the system complete the program in multiple steps. The following two considerations motivate this approach.

First, we speculate that factorizing a potentially long and complicated specification into multiple steps would ease the understanding by a model and hence enhance program synthesis. In the multi-turn approach, a model can focus on the specification associated with one subprogram and avoid arduously tracking the complicated dependency among subprograms. This effectively reduces the search space besides the convenience of specifying user intent. Indeed, our speculations are confirmed in our experiments with higher quality synthesized programs through the multi-turn approach. 

Second, code exhibits a weak pattern of interleaved natural and programming language, which may be exploitable. Such a pattern is formed by programmers who explain the functionality of a program with comments.
With the language modeling objective, we hypothesize that the interleaving pattern provides a supervision signal for the model to generate programs given natural language descriptions over \textit{multiple} turns. The signal is highly noisy or weak, because only a subset of data would exhibit such a pattern, comments may be inaccurate or uninformative, and some of them may even be placed at an irrelevant position. However, up-scaling the model and data size might overcome such weak supervision, allowing the model to develop multi-turn program synthesis capacity.
This enables user intent to be expressed in multiple turns, that is, the intent can be decomposed and fulfilled part by part while each turn can easily be expressed in natural language.

In this work, we develop a multi-turn programming benchmark to measure the models' capacity for multi-turn program synthesis. To solve a problem in the benchmark,
a model needs to synthesize a program in multiple steps with a user who specifies the intent in each turn in natural language. Please refer to Figure~\ref{fig:problem-example} for an example where the model synthesizes a program to extract the user name of an email address. Performance on the benchmark is measured by pass rate on expert-written test cases. To the best of our knowledge, this is the first multi-turn program synthesis benchmark, which allows quantitative analysis of multi-turn program synthesis. With the emergence of multi-turn program synthesis capacity in large language models that benefits problem-solving, we believe this benchmark will foster future research in program synthesis.

\textbf{Our Contributions} 
Our work shares the basic idea of adopting language models for program synthesis with the recent and concurrent efforts~\citep{chen2021evaluating,austin2021program,alphacode} with a single-turn user intent specification. In addition, we contribute with respect to four aspects:

\begin{itemize}[leftmargin=*]
    \itemsep0em 
    \item We study multi-turn program synthesis emerging in autoregressive models under scaling laws.
    \item We leverage this capacity to introduce a multi-turn program synthesis paradigm.
    \item We investigate its properties quantitatively with a novel multi-turn programming benchmark.\footnote{Benchmark: \url{https://github.com/salesforce/CodeGen/tree/main/benchmark}}
    \item We will open source model checkpoints\footnote{Checkpoints: \url{https://github.com/salesforce/CodeGen}} and the custom training library: \jf.\footnote{Training: \url{https://github.com/salesforce/jaxformer}}
\end{itemize}

For program synthesis, no large-scale models competitive with Codex are available as open-source. This hinders progress, given that the expensive compute resources required to train these models are only accessible to a limited number of institutions. Our open source contribution allows a wide range of researchers to study and advance these models, which may greatly facilitate research progress.

\section{Model Training}

To evaluate the emergence of multi-turn programming capabilities under scaling laws, we adopt standard transformer-based autoregressive language models, varying (1) the number of model parameters (350M, 2.7B, 6.1B, 16.1B) and (2) the number of tokens of programming languages in the training corpora. For scaling the training, a custom library \jf for TPU-v4 hardware was developed and will be released as open-source, including the trained model weights.

\subsection{Datasets}
\label{sec:data}

The family of \model models is trained sequentially on three datasets: \dnl, \dpl, and \dmpl.

The natural language dataset \dnl is an $825.18$ GiB English text corpus collected by \cite{gao2020pile} for language modeling (MIT license). The dataset is constructed from 22 diverse high-quality subsets, one of which is programming language data collected from GitHub repositories with >100 stars that constitute 7.6\% of the dataset.
Since the majority of \dnl is English text, the resulting models are called as natural language \model models (\nl).

The multi-lingual dataset \dpl is a subset of Google's publicly available BigQuery dataset, which consists of code (under open-source license) in multiple programming languages. For the multi-lingual training, the following 6 programming languages are chosen: C, C++, Go, Java, JavaScript, and Python. Thus, we refer to models trained on the \dpl as multi-lingual \model models (\pl).

\legal{The mono-lingual dataset \dmpl contains a large amount of data in the programming language, Python. We have compiled public, non-personal information from GitHub consisting of permissively licensed Python code in October 2021.} Consequently, we refer to models trained on \dmpl as mono-lingual \model models (\mpl).

The pre-processing follows: (1) filtering, (2) deduplication, (3) tokenization, (4) shuffling, and (5) concatenation. For details on \dnl, we refer to \cite{gao2020pile}. For \dpl and \dmpl, we refer to Appendix~\ref{app:training}. Table~\ref{table:data} summarizes the statistics of the training corpora.

\subsection{Models}


The \model models are in the form of autoregressive transformers with next-token prediction language modeling as the learning objective trained on a natural language corpus and programming language data curated from GitHub. The models are trained in various sizes with 350M, 2.7B, 6.1B, and 16.1B parameters. The first three configurations allow for direct comparison with open-sourced large language models trained on text corpus, \gptn (350M, 2.7B)~\citep{gpt-neo} and \gptj (6B)~\citep{gpt-j}. See Table~\ref{table:training} in Appendix~\ref{app:training} for model specifications.

The \model models are trained in a sequential nature over datasets. \nl is first trained on \dnl. \pl is initialized from \nl and trained on \dpl. Finally \mpl is initialized from \pl and trained on \dmpl.

The emergence of program synthesis conditional on descriptions in natural language may stem from the size of the models and data, training objective, and nature of the training data itself. This is called emergence since we do not explicitly train the model on comment-code pairs. Similar phenomena are observed in a wide range of natural language tasks where a large-scale unsupervised language model can solve unseen tasks in a zero-shot fashion~\citep{brown2020language}. The emergence phenomena or surprising zero-shot generalization is often attributed to the large scale of the model and the data.

While our focus is not to reveal the underlying mechanism on why program synthesis capacity emerges from simple language modeling, we make an attempt to provide an explanation given the nature of our modeling approach and the training data. The data consists of regular code from GitHub (without manual selection), for which \textit{some} data exhibits a pattern of interleaved natural and programming language, which we believe provides a noisy supervision signal for the program synthesis capacity due to the next-token prediction training objective. However, we emphasize that such a data pattern is highly noisy and weak, because only a subset of data exhibits such a pattern, e.g., comments may be inaccurate or uninformative, and some of them may even be placed at an irrelevant position. Therefore, we believe two main factors contribute to the program synthesis capacity: 1) large scale of model size and data size and 2) noisy signal in training data.

The scaling of such LLMs requires data and model parallelism. To address these requirements, a training library \jf (\url{https://github.com/salesforce/jaxformer}) was developed for efficient training on Google's TPU-v4 hardware. We refer to Appendix~\ref{app:training} for further details on the technical implementation and sharding schemes. Table~\ref{table:training} summarizes the hyper-parameters.

\section{Single-Turn Evaluation}
\label{sec:single-turn}

We first evaluate our \model using an existing program synthesis benchmark: HumanEval (MIT license)~\citep{chen2021evaluating}.
HumanEval contains $164$ hand-written Python programming problems. Each problem provides a prompt with descriptions of the function to be generated, function signature, and example test cases in the form of assertions. The model needs to complete a function given the prompt such that it can pass all provided test cases, thus measuring the performance by functional correctness. Since a user intent is specified in a single prompt and provided to the model once, we regard the evaluation on HumanEval as a single-turn evaluation, to distinguish it from the multi-turn evaluation which we introduce in the next section. Following \cite{chen2021evaluating}, we recruit nucleus sampling~\citep{holtzman2019curious} with top-$p$ where $p = 0.95$. 

\begin{table}[tb]
\begin{center}
\scalebox{\scale}{
\begin{footnotesize}
\begin{tabular}{p{6cm}ccc}
\toprule
\multirow{2}{*}{Model} & \multicolumn{3}{c}{pass@$k$ {[\%]}} \\\cmidrule{2-4}
& $k=1$ & $k=10$ & $k=100$ \\
\midrule
\gptnt & \phantom{0}0.85 &  \phantom{0}2.55 & \phantom{0}5.95 \\
\gptns & \phantom{0}6.41 & 11.27 & 21.37 \\
\gptj 6B & 11.62 & 15.74 & 27.74 \\
\midrule
\codext & 13.17 & 20.37 & 36.27 \\
\codexs & 21.36 & 35.42 & 59.50 \\
\codexl & 28.81 & 46.81 & 72.31 \\
code-cushman-001$^\ast$ & 33.5\phantom{0} & 54.3\phantom{0} & 77.4\phantom{0} \\
code-davinci-001$^\ast$ & 39.0\phantom{0} & 60.6\phantom{0} & 84.1\phantom{0} \\
code-davinci-002$^\ast$ & 47.0\phantom{0} & 74.9\phantom{0} & 92.1\phantom{0} \\
\midrule
\nlt & \phantom{0}2.12 & \phantom{0}4.10 & \phantom{0}7.38 \\
\nls & \phantom{0}6.70 &  14.15 & 22.84 \\
\nll & 10.43 & 18.36 & 29.85 \\
\nlxl & 14.24 & 23.46 & 38.33 \\
\midrule
\plt & \phantom{0}6.67 & 10.61 & 16.84 \\
\pls & 14.51 & 24.67 & 38.56 \\
\pll & 18.16 & 28.71 & 44.85 \\
\plxl & 18.32 & 32.07 & 50.80 \\
\midrule
\mplt & 12.76 & 23.11 & 35.19 \\
\mpls & 23.70 & 36.64 & 57.01 \\
\mpll & 26.13 & 42.29 & 65.82 \\
\mplxl & \textbf{29.28} & \textbf{49.86} & \textbf{75.00}\\
\bottomrule
\end{tabular}
\end{footnotesize}
}
\end{center}
\caption{
Evaluation results on the HumanEval benchmark. Each pass@$k$ (where $k \in \{1, 10, 100\}$) for each model is computed with three sampling temperatures ($t \in \{0.2, 0.6, 0.8$\}) and the highest one among the three are displayed, which follows the evaluation procedure in \cite{chen2021evaluating}. Results for the model marked with $^\ast$ are from \cite{chen2022codet}.
}
\label{tab:humaneval}
\end{table}


\subsection{HumanEval Performance Scales as a Function of Model Size and Data Size}
\label{sec:humaneval-scale}

We compare our models to the Codex models~\citep{chen2021evaluating}, which demonstrate the state-of-the-art performance on HumanEval. Moreover, our models are compared to open-sourced large language models, \gptn~\citep{gpt-neo} and \gptj~\citep{gpt-j}. These are trained on \dnl~\citep{gao2020pile}, and thus similar to our \nl models, in terms of training data and model size. All models are evaluated with temperature  $t \in \{0.2, 0.6, 0.8\}$, and we compute pass@$k$ where $k \in \{1, 10, 100\}$ for each model. For direct comparison to the results by \cite{chen2021evaluating}, we choose the temperature that yields the best-performing pass@$k$ for each $k$. The results of our models and baselines are summarized in Table~\ref{tab:humaneval}. Our \nl models (350M, 2.7B, 6.1B) outperform or perform on par with the respective \gptn and \gptj models. 
Further training \nl on multilingual programming language data (\dpl) leads to \pl.
The multilingual \model models outperform the models trained on \dnl (\gptn, \gptj, \nl) by a large margin. 
We then finetune \pl on a Python-only dataset (\dmpl), resulting in \mpl. The program synthesis capacity is improved substantially. 
Therefore, the Python program synthesis capacity enhances as the amount of Python training data increases. For almost all models, as expected, increasing the size of the model improves overall performance.

Our Python-monolingual \model models have competitive or improved performance, compared to the current state-of-the-art models, Codex. \mpls underperforms \codexs when $k=100$ but outperforms it when $k \in \{1, 10\}$. While it is only half the size, our \mpll demonstrates pass@k scores approaching those of the best-performing Codex, \codexl. Our largest model \mplxl is competitive or outperforms it depending on $k$.

\subsection{Better User Intent Understanding Yields Better Synthesized Programs}
\label{sec:humaneval-ppl}

The success of a program synthesis system highly depends on how well it understands user intent. When the system is based on a language model, the perplexity of problem prompts provides a proxy for the system's understanding of user intent specifications. A low perplexity of an intent specification under a model indicates that this intent specification is compatible with the knowledge learned by the model from the training data. We investigate whether better prompt understanding, with lower prompt perplexity as a proxy, leads to more functionally accurate programs. 



\begin{table}[t]
\begin{center}
\scalebox{\scale}{
\begin{footnotesize}
\begin{tabular}{p{3cm}cccc}
\toprule
\mpl & 350M & 2.7B & 6.1B & 16.1B  \\
\midrule
Pass & $3.78\pm0.23$ & $3.66\pm0.14$ & $3.35\pm0.13$ & $3.12\pm0.11$ \\
Non-Pass & $5.18\pm0.19$ & $4.37\pm0.18$ & $3.88\pm0.13$ & $3.40\pm0.11$\\
\bottomrule
\end{tabular}
\end{footnotesize}
}
\end{center}
\caption{
Average prompt perplexity$^\downarrow$ ($\pm$ standard error) of \mpl models on pass and non-pass problems.}
\label{tab:humaneval-ppl-mono}
%
\end{table}

We partition all problems into pass versus non-pass ones. A pass problem is one that at least one sample from 200 samples passes all test cases, while for a non-pass problem none of the 200 samples pass all test cases. We compute the average perplexity of the problem prompts of the pass problems and that of the non-pass ones, based on samples from \mpl models.
The results are displayed in Table~\ref{tab:humaneval-ppl-mono} (see Appendix~\ref{ref:ppl-nl-and-multi} for the results on \nl and \pl). The prompts of the pass problems have lower perplexity than those of the non-pass ones. This finding implies that program synthesis is more likely to be successful when the user intent specification is understood better by the model.
Indeed, some training data contains interleaved sequences of natural language comments and programs, where the comments  describe the functionality of the following program.
We thus speculate that user intent specifications similar to such a pattern would be better understood by the model, and hence lead to better program synthesis. Inspired by this pattern, we propose to specify user intent in multiple turns such that the model focus on a partial problem at a time, which would make user intent understanding by the model easier. 



\section{Multi-Turn Evaluation}


In this section, we propose and study a multi-step program synthesis paradigm where program synthesis is decomposed into multiple steps and the system synthesizes a subprogram in each step. To examine such a paradigm, we first develop a \mtpbf (\mtpb). \mtpb consists of $115$ problems written by experts, each of which includes a \textit{multi-step} descriptions in natural language (\textit{prompt}). To solve a problem, a model needs to synthesize 
functionally correct subprograms (1) following the description at the current step and (2) considering descriptions and synthesized subprograms at previous steps (\textit{e.g.}, correct backreference of functions and/or variables defined in the previous steps). An illustrative example is shown in Figure~\ref{fig:problem-example}.

\subsection{Benchmark Construction}



We (4 authors) start by defining\footnote{Problem writing was performed in a closed book format, \textit{i.e.} we are not allowed to consult with online resources while writing the problems.} a set of $115$ problems requiring a diverse range of programming knowledge, including math, array operations, string manipulations, algorithms, data science, and problems that require other knowledge, such that the number of problems in each category is roughly balanced.\footnote{See Appendix~\ref{app:prob_list} for a complete listing.}
For each problem, we construct a triplet consisting of multi-turn prompts $P$, test case inputs $I$, and test case outputs $O$. 
Multi-turn prompts $P$ are designed following the two constraints: (1) the problem is decomposed into 3 or more turns, (2) a single turn cannot be attributed to solving the problem. For example, implementing a linear regression model could be phrased as ``Perform linear regression on x and y''. Since the main task is fully expressed in this prompt, understanding this prompt is sufficient to perform the task. We avoid such cases via manual inspection and distribute problem-solving over turns.
Together with the prompts, we task the problem author to prepare 5 sets of test case inputs $I$ and outputs $O$ to evaluate model outputs  with functional correctness.
To reduce wrongly rewarding false positive solutions that give meaningless programs but pass the tests, we examine and revise such cases to ensure the test quality.

\begin{figure*}[t]
	\centering	
	\includegraphics[width=0.75\linewidth]{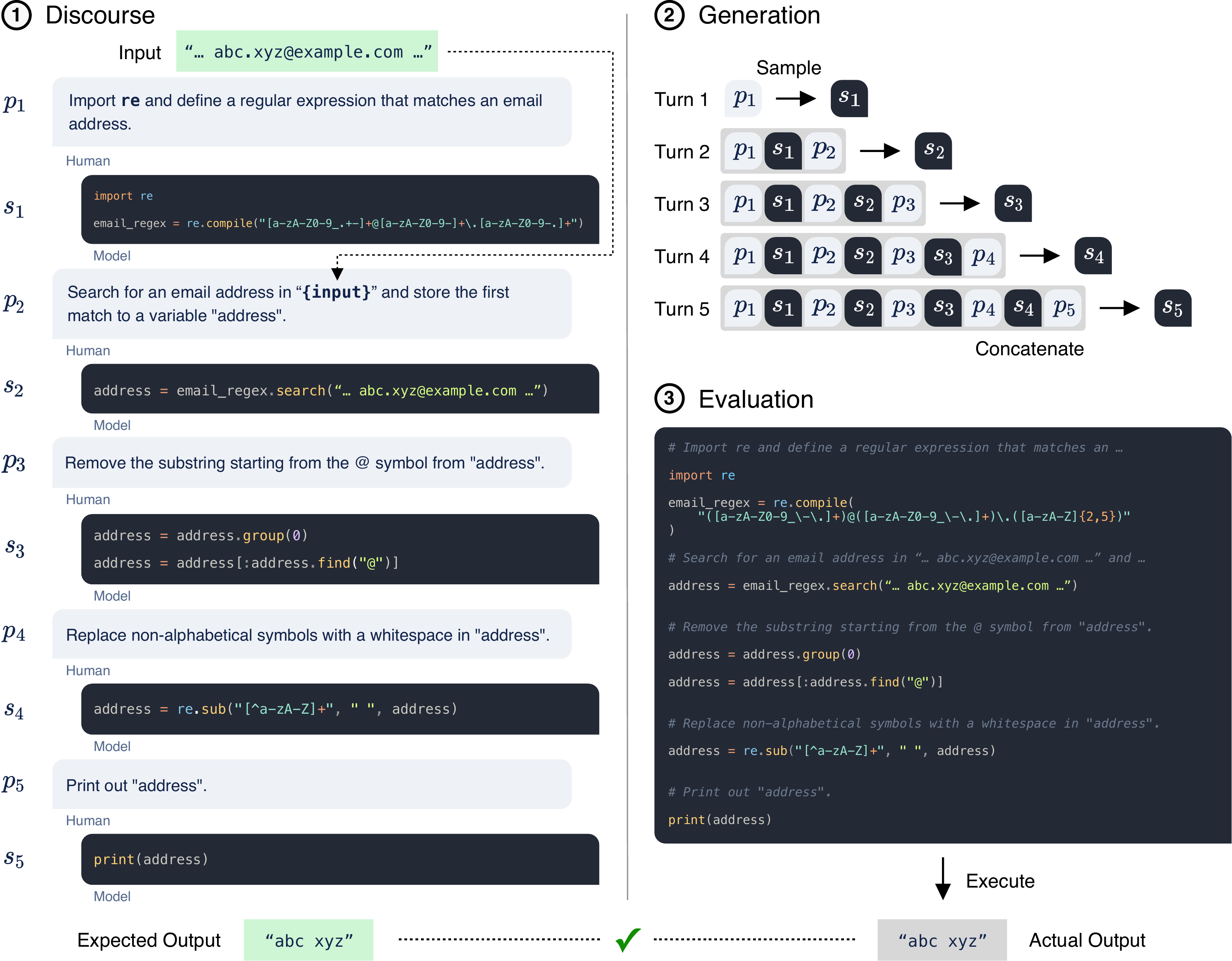}
	\caption{An illustrative example for the \mtpbf, performing the task of extracting the user name of an email address. \circled{1} Each problem consists of prompts $p_i$ and unit tests, where some prompts include templates (\textit{i.e.} \texttt{\{input\}}) that are filled with test case inputs before it is fed to the model. In the displayed example, the input is a string containing \texttt{abc.xyz@example.com}, which replaces \texttt{\{input\}} in \texttt{$p_2$}, and the expected output is \texttt{abc xyz}. \circled{2} Our model conditions on the concatenation of interleaved past prompts and \textit{generated responses}. \circled{3} Generated responses from each turn are concatenated and executed, where the output is compared to the answer.}
	\label{fig:problem-example}
 \vspace{-2mm}
\end{figure*}


Unlike HumanEval for which models are expected to complete a partially defined function, \mtpb problems only provide the prompts, thereby models have to generate the solution from scratch.\footnote{To guide sampling in Python, we prefix the prompt with: \texttt{\# Import libraries.{\textbackslash}n import numpy as np}.}
While the free-form generation may allow for more potential solutions, the lack of an entry point to provide test case inputs makes it challenging to test the generated code on diverse test cases.
To overcome this challenge, we instead embed test case inputs within prompts.
Specifically, prompts are written with Python's formatted string\footnote{\url{https://docs.python.org/3/reference/lexical_analysis.html##f-strings}} where input values are substituted for the variable name when a specific test case is applied to the problem.
For example, a prompt, ``Define a string named `s' with the value \{\texttt{var}\}.'',  together with a test case input \texttt{var = `Hello'} will be formatted into ``Define a string named `s' with the value `Hello'.'' Also see \circled{1} in Figure~\ref{fig:problem-example} for an example.

\subsection{Execution Environment and Solution Evaluation}




For execution, the history of pairs of prompts and generated completions is concatenated into a self-contained program (see \circled{3} in Figure~\ref{fig:problem-example} for an example). The program is then executed in an isolated Python environment following the single-turn HumanEval benchmark \citep{chen2021evaluating}. However, the problems in HumanEval are constructed in such a way that a known function signature is completed, thus invocation of the generated code under a set of functional unit tests is trivial. In our multi-turn case, no such entry point (or return value) is guaranteed to be generated. To circumvent the issue of a missing return signature (or value), the last prompt of the multi-turn problems in \mtpb is always specified to print out the resulting state to the terminal. Then, the benchmark execution environment overloads the Python \texttt{print(args)} function and stores \texttt{args} on a stack. If the sampled code for the last prompt of a problem does not include the \texttt{print()} statement, which is a valid convention to print on the terminal in Python or specifically Jupyter notebooks, then the AST of the generated code will be mutated to inject an invocation of \texttt{print()}. Finally, a type-relaxed equivalence check (\textit{e.g.}, an implicit conversion between lists and tuples) of \texttt{args} against the predefined gold output of the problem is performed to determine test failure or success. 

\begin{table}[tb]
\begin{center}
\scalebox{\scale}{
\begin{footnotesize}
\begin{tabular}{llccccc}
\toprule
\multirow{2}{*}{Data} &\multirow{2}{*}{Model} & \multicolumn{4}{c}{Pass Rate$^\uparrow$ {[\%]}}\\\cmidrule{3-7}
& & 350M & 2.7B & 6.1B & 16.1B & - \\
\midrule
\dnl & \gptn \& \gptj & \phantom{0}0.79 & \phantom{0}8.17 & 18.86 & - & -\\
\dnl & \nl & \phantom{0}0.23 & 15.31 & 19.37 & 30.33 & -\\
\dpl & \pl  & \phantom{0}4.09 & 20.82 & 25.51 & 26.27 & -\\
\dmpl & \mpl  & 16.98 & 38.72 & 43.52 & 47.34 & -\\
- & code-cushman-001  & -     & -     & -     & -   & 56.77 \\
- & code-davinci-001  & -     & -     & -     & -   & 55.28 \\
- & code-davinci-002  & -     & -     & -     & -   & 59.86 \\
\bottomrule
\end{tabular}
\,
\end{footnotesize}
}
\end{center}
\vspace{-1mm}
\caption{
Evaluation results on the \mtpbf. The multi-turn program synthesis performance varies as a function of model size (columns) and code data size (rows).
}
\vspace{-1mm}
\label{tab:multiturn-eval}
\end{table}

\begin{table}[tb]
\begin{center}
\scalebox{\scale}{
\begin{footnotesize}
\noindent\makebox[\textwidth]{%
\begin{tabular}{p{1.55cm}ccccp{0pt}cccc}
\toprule
\multirow{2}{*}{Prompt} & \multicolumn{4}{c}{PPL$^\downarrow$} & & \multicolumn{4}{c}{Pass Rate$^\uparrow$ {[\%]}}\\\cmidrule{2-5}\cmidrule{7-10}
& 350M & 2.7B & 6.1B & 16.1B & & 350M & 2.7B & 6.1B & 16.1B \\\midrule
Single-Turn & $13.92\pm1.89$  & $11.67\pm1.46$ & $10.58\pm1.20$ & $10.25\pm0.99$ & & \phantom{0}5.75 & 25.43 & 28.48 & 38.74\\
Multi-Turn & $10.09\pm0.62$ & \phantom{0}$8.90\pm0.52$ & \phantom{0}$8.18\pm0.43$ & \phantom{0}$8.05\pm0.43$  & & 16.98 & 38.72 & 43.52 & 47.34\\
\bottomrule
\end{tabular}
}
\end{footnotesize}
}
\end{center}
\caption{
Comparison between multi- and concatenated single-turn specifications on perplexity (PPL) and program synthesis performance (as measured by pass rate) under \mpl models.
}
\vspace{-2mm}
\label{tab:multi_single_comp}
\end{table}

\subsection{Multi-Step Programming Capacity Scales with Model Size and Data Size}
In this analysis, we investigate how the model size and data size affect the program synthesis capacity in a multi-turn paradigm. In the \mtpb, each problem has 5 test cases and we sample 40 samples for each test case with each model, based on which the pass rate is computed for each problem. The \mtpb evaluation results (average pass rate) for our \model models, baselines, and OpenAI Codex models\footnote{Accessed on November 10th, 2022.} are shown in Table~\ref{tab:multiturn-eval}. Clearly, the performance on the \mtpb improves as a function of the model size and data size. This suggests that the capacity of multi-step program synthesis scales as a function of the model size and data size. The models are simply trained with an autoregressive language modeling objective. While the model and the data scale up, multi-turn program synthesis capacity emerges, that is, the capacity to synthesize programs in a multi-turn fashion.

\subsection{Better User Specification Understanding with Multi-turn Factorization}
We hypothesize that multi-turn factorization enhances the model's understanding of user intent specifications, which in turn lead to higher program synthesis capacity. To test this hypothesis, we form a single-turn counterpart of multi-turn specifications by concatenating each specification into a single turn. As discussed in Section~\ref{sec:humaneval-ppl}, we adopt the prompt perplexity as a proxy for user intent understanding. Thus, we compare the perplexity of the multi-turn prompts and that of the concatenated single-turn prompts under the four \mpl models.

The average perplexity (see Appendix~\ref{app:ppl} for the calculation details) over all the problems in the \mtpb is displayed in the left panel of Table~\ref{tab:multi_single_comp}. For all models, the single-turn specification has a higher average perplexity than the multi-turn specification. It implies that the multi-turn user specifications can be better understood by the models. We notice that the average perplexity for both multi-turn and single-turn intent specifications under larger models is slightly lower than that under smaller models, indicating that the larger ones understand the user intent better than the smaller ones.  

We compare the program synthesis pass rate with the multi-turn prompts to that with the concatenated single-turn prompts. The results are shown in the right panel of Table~\ref{tab:multi_single_comp}. Multi-turn specifications lead to close to or more than 10 percentage points over single-turn specifications for all model sizes. Together with the perplexity analysis above, it appears that factorizing a user specification into multiple steps and leveraging the emerged capacity of large language models allow them to digest the specification more easily and synthesize programs more successfully.

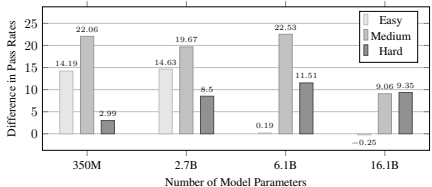
\begin{figure}[tb]
	\centering	
	\scalebox{0.5}{%
        \begin{tikzpicture}
            \begin{axis}[
                height=5.4cm,
                width=0.85\linewidth,
                ymajorgrids = true,
                ybar=.2cm,
                ymax=25,
                ymin=0.0,
                ytick={0.0,5,10,...,100},
                yticklabel style={
                    /pgf/number format/fixed,
                    /pgf/number format/precision=2
                },
                enlargelimits=0.15,
                legend style={font=\footnotesize},
                legend pos=north east,
                ylabel={Difference in Pass Rates},
                xlabel={Number of Model Parameters},
                label style={font=\footnotesize},
                xticklabel style={font=\footnotesize, rotate=0},
                symbolic x coords={350M,2.7B,6.1B,16.1B},
                xtick=data,
                nodes near coords,
                nodes near coords style={font=\tiny}]
                \addplot[fill=gray!20,draw=gray!50,fill opacity=0.95] coordinates {(350M,14.19) (2.7B,14.63) (6.1B,0.19) (16.1B,-0.25)};
                \addplot[fill=gray!50,draw=gray!80,fill opacity=0.95] coordinates {(350M,22.06) (2.7B,19.67) (6.1B,22.53) (16.1B,9.06)};
                \addplot[fill=gray!90,draw=black!70,fill opacity=0.95] coordinates {(350M,2.99) (2.7B,8.50) (6.1B,11.51) (16.1B,9.35)};
                \legend{Easy,Medium,Hard}
            \end{axis}
        \end{tikzpicture}
    }
	\caption{Difference in average pass-rate of problems in single-turn and multi-turn formulation over levels of problem difficulty. The improvement is sizable for most model sizes and difficulty levels, except for easy problems with larger models.}
	\label{fig:multi-turn-improvement}
	
\end{figure}

Furthermore, we categorize the problems by difficulty level based on their average pass rates (``hard'' with less than 30\%, ``easy'' with larger than 70\%), and examine the interaction effect between difficulty level and model size on the improvement by multi-turn factorization. See the results in Figure~\ref{fig:multi-turn-improvement}. Across almost all model sizes and difficulty levels, multi-turn prompts lead to significant improvement over single-turn prompts and most improvements are nearly or higher than $10$ percentage points. Interestingly, the larger models (6.1B and 16.1B) are invariant to multi-turn factorization for easy problems (see the two short bars, $0.19\%$ and $-0.25\%$, in Figure~\ref{fig:multi-turn-improvement}). This implies that when the problems can be easily understood by the model (due to the combined effect of easiness of the problems and the high capacity of larger models), it is not necessary or beneficial to factorize the specifications. This is in fact consistent with our motivating assumption that factorizing complicated specifications would ease problem understanding and improve program synthesis.   

			
			

\subsection{Qualitative Examples}
\label{sec:qualitative_example}

To further understand the differences in model behavior over model sizes, we examine cases where large models have contrasting performances to smaller models. We specifically select problems for which \mplxl and \mpls show a significant discrepancy in performance.
On problems where \mplxl performed significantly worse compared to \mpls, we observe that the larger model becomes inflexible due to taking the prompt literally.
For example, initializing a number always results in an integer, despite the prompt asking to cast into a string~(Figure \ref{lst:problem_96}), or the ``return'' keyword in a prompt triggers a function definition while the intent is to directly generate an executable program~(Figure \ref{lst:problem_62}).
However in general, larger-scale models overcome mistakes due to prompt misinterpretation by smaller models, including assigning multiple variables at the same time (Figure \ref{lst:problem_95}) or understanding the concept of \texttt{any} comparison (Figure \ref{lst:problem_48}).



\section{Related Work}

\paragraph{Program Synthesis} While program synthesis has a long history, two inherent challenges remain unsolved: (1) intractability of the program space and (2) difficulty in accurately expressing user intent \citep{manna1971toward, gulwani2017program}. A large body of prior research attempted to address (1) by exploring methods like stochastic search techniques~\citep{parisotto2016neuro, schkufza2013stochastic} and deductive top-down search \citep{gulwani2011automating, polozov2015flashmeta}. However, the scalability of these approaches is still limited. User intent can be expressed with various methods: formal logical specifications, input-output examples, and natural language descriptions. Complete and formal specifications require too much effort, while informal ones like input-output examples often under-specify problems~\citep{gulwani2011automating}. Well-learned conditional distribution and language understanding capacity owing to the large-scale model and data allows for efficient solutions for these two challenges.   
Several works investigate converting conversational intents into programmable representations, such as SQL~\citep{yu-etal-2019-cosql,yu-etal-2019-sparc} or dataflow graph~\citep{andreas2020task}. Our proposed benchmark requires the generation of Python, which is more general and complex.

\paragraph{Large Language Models}
Transformers capture dependency among sequence elements through attention mechanism \citep{bahdanau2014neural} and are highly scalable. It has been successfully applied to natural language processing \citep{devlin2018bert, lewis2020bart, raffel2020exploring}, computer vision \citep{dosovitskiy2020image}, and many other areas \citep{oord2018representation, jumper2021highly}. Prior works, such as CuBERT \citep{kanade2020learning}, CodeBERT \citep{feng2020codebert}, PyMT5 \citep{clement2020pymt5}, and CodeT5 \citep{wang2021codet5}, have applied transformers towards code understanding but these mostly focus on code retrieval, classification, and program repair. Several recent and concurrent efforts explore using large language models for program synthesis~\citep{chen2021evaluating, austin2021program, alphacode,fried2022incoder} and its effectiveness~\citep{vaithilingam2022expectation}. While they focus on generating code in a single turn, we propose to factorize the specifications into multiple turns and demonstrate that it is highly effective to improve synthesis quality. It is worth pointing out that \cite{austin2021program} explored refining the code in multiple iterations, but it is essentially a single-turn approach since a complete program is produced in every single turn.
Prompting pre-trained language models with intermediate information to improve task performance has attracted interest~\citep{nye2021show,wei2022chain}. Our proposed MTPB also allows the model to leverage past turns as context.


\paragraph{Benchmarks for Program Synthesis}

To quantitatively evaluate program synthesis models, several benchmarks have been proposed with different input forms. 
A popular input forms include preceding code in the same line~\citep{raychev2016probabilistic}, pseudo-code~\citep{kulal2019spoc}, a docstring and function signature~\citep{chen2021evaluating}, or problem description~\citep{hendrycks2021measuring}.
In most of those cases, only directly relevant input information is given to the model.
In contrast, a few previous works instantiate benchmarks that measure the ability to generate programs given surrounding program context beyond the target program, such as variables and other methods~\citep{iyer-etal-2018-mapping} or alternating ``cells'' of preceding code and text blocks~\citep{agashe2019juice}, while the primary focus is to generate the target program itself.
We propose a new benchmark that requires a progressive generation of subprograms through multi-turn prompts.


\section{Conclusion}
We study program synthesis with large causal language models trained on large corpora of code data. The capacity to understand long context and generate coherent responses emerges from the simple language modeling as the model size and data size scale up. Leveraging this capacity and observing that better user intent understanding leads to better program synthesis, we propose a multi-step program synthesis approach in which program synthesis is achieved through a multi-turn specification and code generation. Moreover, we develop the \mtpbf (\mtpb) to investigate our models' capacity on synthesizing programs in such a multi-step paradigm. Our experiments show that the multi-step program synthesis capacity scales as a function of the model size and data size. The intent specifications, which are specified in multiple steps, are digested more easily by the models and lead to more accurate program synthesis. We open-source the training code and the model checkpoints to facilitate future research and practical applications in this area. 

\newpage

\section*{Broader Impact and Ethical Considerations}
\label{app:impact}
All variants of \model are firstly pre-trained on the Pile, which includes a small portion of profane language. Focusing on the GitHub data that best aligns our expected use case of program synthesis, \cite{gao2020pile} report that 0.1\% of the data contained profane language, and has sentiment biases against gender and certain religious groups.
Thus, while we did not observe in our samples, \model may generate such content as well.
In addition to risks on natural language outputs (\textit{e.g.}, docstrings), generated programs may include vulnerabilities and safety concerns, which are not remedied in this work.
Models should not be used in applications until being treated for these risks.

\bibliography{anthology,custom}
\bibliographystyle{iclr2023_conference}

\clearpage
\onecolumn
\appendix

\section{Model Training}
\label{app:training}

To evaluate the emergence of multi-turn program synthesis capabilities under scaling laws, we adopt standard transformer-based autoregressive language models, varying (1) the number of model parameters (350M, 2.7B, 6.1B, 16.1B) and (2) the number of tokens of programming languages in the training corpora. For scaling the models, a custom library \jf for training large language models on TPU-v4 hardware was developed and will be released as open source, including the trained model weights.

\subsection{Datasets}
\label{app:data}


\begin{table*}[th]
\begin{center}
\begin{small}
\begin{tabular}{llrrrr}
\toprule
Dataset & Language & Raw Size & Final Size & Final Tokens\\
\midrule
\multirow{2}{*}{\dnl} & Natural Language & $825.18$ GiB & $1159.04$ GiB & $354.7 $B \\
& Code & $95.16$ GiB & $95.16$ GiB & $31.6$B \\
\midrule
\multirow{6}{*}{\dpl} & C & $1772.1$ GiB & $48.9$ GiB & $19.7$B \\
& C++ & $205.5$ GiB & $69.9$ GiB & $25.5$B \\
& Go & $256.4$ GiB & $21.4$ GiB & $9.6$B \\
& Java & $335.1$ GiB & $120.3$ GiB & $35.4$B \\
& JavaScript & $1282.3$ GiB & $24.7$ GiB & $9.7$B \\
& Python & $196.8$ GiB & $55.9$ GiB & $19.3$B \\
\midrule
\multirow{1}{*}{\dmpl} & Python & $5558.1$ GiB & $217.3$ GiB & $71.7$B \\
\bottomrule
\end{tabular}
\end{small}
\end{center}
\caption{Approximate statistics for training corpora along the pre-processing steps.}
\label{table:data}
\end{table*}





For each dataset, the pre-processing shares the following steps: (1) filtering, (2) deduplication, (3) tokenization, (4) shuffling, and (5) concatenation. For details on \dnl, we refer to \cite{gao2020pile}. For \dpl and \dmpl, in (1) files are filtered by file extension, and files with average lines length of <100 characters, a maximum line length of $1,000$, and >90\% of the characters being decimal or hexadecimal digits are removed. For (2), exact duplicates based on their SHA-256 hash are removed, which amounts to a substantial portion of the raw data due to forks and copies of repositories. For (3), the BPE vocabulary of GPT-2 is extended by special tokens representing repeating tokens of tabs and white spaces. In the multi-lingual setting of \dpl, a prefix is prepended to indicate the name of the programming language. For (4), each year of data is randomly shuffled. For (5), sequences are concatenated to fill the context length of $2,048$ tokens with a special token as a separator. Table~\ref{table:data} summarizes the statistics of the training corpora.

\nl models are randomly initialized and trained on \dnl. \pl models are initialized from \nl and then trained on the \dpl. \mpl models are initialized from \pl and then trained on \dmpl.    





\begin{table}[t]
\begin{center}
\begin{small}
\begin{tabular}{lllrrrr}
\toprule
Model & Dataset & Hyper-parameter & 350M & 2.7B & 6.1B & 16.1B\\
\midrule
\multirow{6}{*}{\model} & & Number of layers & 20 & 32 & 33 & 34 \\
& & Number of heads & 16 & 32 & 16 & 24 \\
& & Dimensions per head & 64 & 80 & 256 & 256 \\
& & Context length & 2,048 & 2,048 & 2,048 & 2,048 \\
& & Batch size & 500k & 1M & 2M & 2M \\
& & Weight decay & 0.1 & 0.1 & 0.1 & 0.1 \\
\midrule
\multirow{3}{*}{\nl} & \multirow{3}{*}{\dnl} &  Learning rate & $3.0\mathrm{e}{-4}$ & $1.6\mathrm{e}{-4}$ & $1.2\mathrm{e}{-4}$ & $0.9\mathrm{e}{-4}$ \\
& & Warm-up steps & 3k & 3k & 3k & 3k \\
& & Warm-up / Total steps & 350k & 350k & 350k & 350k \\
\midrule
\multirow{3}{*}{\pl} & \multirow{3}{*}{\dpl} & Learning rate & $1.8\mathrm{e}{-4}$ & $0.8\mathrm{e}{-4}$ & $0.4\mathrm{e}{-4}$ & $0.5\mathrm{e}{-4}$ \\
& & Warm-up steps & 3k & 3k & 3k & 3k \\
& & Total steps & 150k & 150k & 150k & 150k \\
\midrule
\multirow{3}{*}{\mpl} & \multirow{3}{*}{\dmpl} & Learning rate & $1.8\mathrm{e}{-4}$ & $0.8\mathrm{e}{-4}$ & $0.4\mathrm{e}{-4}$ & $0.5\mathrm{e}{-4}$ \\
& & Warm-up steps & 3k & 3k & 3k & 3k \\
& & Total steps & 150k & 150k & 150k & 150k \\
\bottomrule
\end{tabular}
\end{small}
\end{center}
\caption{Hyper-parameters for model specification and optimization for the family of \model models.}
\label{table:training}
\end{table}


\subsection{Models}


Our models are autoregressive transformers with the regular next-token prediction language modeling as the learning objective. The family of \model models is trained in various sizes with 350M, 2.7B, 6.1B, and 16.1B parameters. The first three configurations allow for direct comparison with open-sourced large language models trained on text corpus, \gptn (350M, 2.7B)~\citep{gpt-neo} and \gptj (6B)~\citep{gpt-j}. See Table~\ref{table:training} in Appendix~\ref{app:training} for model specifications.

The architecture follows a standard transformer decoder with left-to-right causal masking. For the positional encoding, we adopt rotary position embedding~\citep{su2021roformer}. For the forward pass, we execute the self-attention and feed-forward circuits in parallel for improved communication overhead following~\cite{gpt-j}, that is, $x_{t+1} = x_t + {\rm mlp}({\rm ln}(x_t + {\rm attn}({\rm ln}(x_t))))$ is altered to $x_{t+1} = x_t + {\rm attn}({\rm ln}(x_t)) + {\rm mlp}({\rm ln}(x_t))$ for which the computation of self-attention, ${\rm attn()}$, and feed-forward, ${\rm mlp()}$, with layer-norm, ${\rm ln()}$, is simultaneous. The architecture and hyper-parameter choices were optimized specifically for the hardware layout of TPU-v4.

\subsection{Training}


The scaling of large language models requires data and model parallelism. Google's TPU-v4 hardware with a high-speed toroidal mesh interconnect naturally allows for efficient parallelism. To efficiently utilize the hardware, the training of the models is implemented in JAX~\citep{jax2018github}. For parallel evaluation in JAX the $pjit()$\footnote{\url{https://jax.readthedocs.io/en/latest/_modules/jax/experimental/pjit.html}} operator is adopted. The operator enables a paradigm named single-program, multiple-data (SPMD) code, which refers to a parallelism technique where the same computation is run on different input data in parallel on different devices.\footnote{\url{https://jax.readthedocs.io/en/latest/jax-101/06-parallelism.html}} Specifically, $pjit()$ is the API exposed for the XLA SPMD partitioner in JAX, which allows a given function to be evaluated in parallel with equivalent semantics over a logical mesh of compute.

Our library \jf recruits a designated coordinator node to orchestrate the cluster of TPU-VMs\footnote{\url{https://cloud.google.com/blog/products/compute/introducing-cloud-tpu-vms}} with a custom TCP/IP protocol. For data parallelism, the coordinator partitions a batch and distributes the partitions to the individual TPU-VMs. For model parallelism, two schemes for the sharding of model parameters are supported
: (1) Intra-TPU-VM, where parameters are sharded across MXU cores\footnote{Specifically, 4 TPU-v4 chips (\textit{i.e.}, 8 physical which amount 4 logical or virtual MXU cores).} inside a physical TPU-v4 board and replicated across boards following \cite{shoeybi2019megatron, gpt-j}; (2) Inter-TPU-VM, where parameters are sharded across TPU-v4 boards and activations are replicated following \cite{rajbhandari2020zero}.

Both intra-TPU-VM and inter-TPU-VM sharding schemes are implemented based on our specific \texttt{pjit()} a logical mesh specification $(r, p, c)$ with $r$ replicas of the parameters, $p$ partitions of the parameters, and $c$ logical cores per board over $n_b$ TPU boards with each $n_c$ logical cores such that $d\times p = n_b$ and $r\times p \times c = n_b \times n_c$.


The intra-TPU-VM scheme is adopted for models of size of less or equal to 6B parameters, the total amount of model and optimizer parameters fit into the combined HBM memory of a single TPU-v4 board. For instance, a TPU-v4-512 slice with $n_b=64$ and $n_c=4$ would be configured as $(r, p, c) = (64, 1, 4)$. That is, the parameters are being replicated across $r=64$ boards with $p=1$ total inter-board partitions and intra-board parallelism across $c=4$ logical chips. In this configuration, the mean gradient is accumulated across boards via \texttt{with\_sharding\_constraint()}, effectively emulating the behavior of the \texttt{xmap()}\footnote{\url{https://jax.readthedocs.io/en/latest/_autosummary/jax.experimental.maps.xmap.html}} operator.

The inter-TPU-VM scheme is adopted for models exceeding the size of 6B parameters for which the model and optimizer parameters have to be sharded across TPU-v4 boards. For instance, a TPU-v4-512 slice with $n_b=64$ and $n_c=4$ would be configured as $(r, p, c) = (1, 64, 4)$. For larger slices such as TPU-v4-1024 with $n_b=128$, one may introduce redundancy in the parameter sharding, \textit{e.g.}, $(r, p, c) = (2, 64, 4)$. In this configuration, the activations are replicated across boards via \texttt{with\_sharding\_constraint()}. Moreover, $(r, p, c)$ allows for backwards compatibility for the logical hardware layout transition from TPU-v3 with $c=8$ to TPU-v4 with $c=4$ by adjusting $p$ without the need for re-sharding.

For the optimization, Table~\ref{table:training} summarizes the hyper-parameters. We adopt the Adam~\citep{kingma2014adam} optimizer with $(\beta_1, \beta_2, \epsilon) = (0.9, 0.999, 1\mathrm{e}{-08})$ and global gradient norm clipping~\citep{pascanu2013difficulty} of $1.0$. The learning rate function over time follows GPT-3~\citep{brown2020language} with warm-up steps and cosine annealing. In summary, we mainly adopted the GPT-3 reference configurations with minor variations accounting for TPU optimizations. We did not have the compute capacity to optimize these hyper-parameters further.

\section{Pass@\texorpdfstring{$k$}{k} Estimator}
\label{app:humaneval-estimator}
We use the unbiased estimator proposed in \cite{chen2021evaluating} to compute pass@$k$. For each task, $n \geq k$ samples are sampled. In particular, we use $n=200$ and $k \leq 100$. Suppose $c$ is the number of correct samples, among the $n$ samples, which pass all the unit tests. Then the unbiased estimator is defined as follows:

\begin{align}
\text{pass@$k$} = \mathbb{E}_{\text{Problems}} \left[ 1 - \frac{{\binom{n-c}{k}}} {\binom{n}{k}} \right]
\end{align}

Directly computing this estimator is numerically unstable. We use the numerically stable numpy implementation introduced by~\cite{chen2021evaluating}.


\section{Type-relaxed Equivalence Check for MTPB Evaluation}
\label{app:check_eq}
We perform the following type-relaxation before assessing the equivalence between model outputs and the expected outputs.
\begin{itemize}
    \item Convert \texttt{numpy} arrays into correspondingly typed lists of standard types (\textit{e.g.} \texttt{np.int32} will be cast to \texttt{int}).
    \item \texttt{pandas} series are converted and compared in \texttt{numpy} array format.
    \item For the rest, model outputs are cast into the type of gold standard outputs.
    \item Floating numbers are compared with $\varepsilon = 1e^{-6}$ as the tolerance threshold.
\end{itemize}

\clearpage
\section{List of \mtpb Problems}
\label{app:prob_list}
\begin{table*}[h!]
\footnotesize
\centering
\begin{tabular}{lll} 
\toprule
 Problem Name              & Problem Description                                                    & Category     \\ \midrule
 Sandwich string           & Append a string in the middle of another string.                       & string       \\
 Normalize integer list    & Normalize a list of positive integers and print formatted percentages. & math         \\
 Convert time              & Convert units of time.                                                 & math         \\
 Squared Fibonacci         & Print the squared Fibonacci numbers.                                   & math         \\
 Compare counts            & Compare the count of positive and negative numbers in a given list.    & array        \\
 Pandas mean               & Construct and compute the mean of a pandas DataFrame.                  & D.S. \\
 Fizz buzz                 & Solve the fizz buzz problem.                                           & Algo.    \\
 Bi-grams                  & Print the bi-grams of a sentence.                                      & string       \\
 Top note                  & Print the name with top note out of a dictionary.                      & dict         \\
 Hex to binary             & Convert hex to binary and reverse.                                     & math         \\
 Invert dict               & Detect an inversion of a given dictionary.                             & dict         \\
 Class definition          & Create a POJO class.                                                   & class        \\
 Longest number            & Print the longest number.                                              & math         \\
 Linear regression         & Fit linear regression model with specified function and sk-learn.      & D.S. \\
 Encrypt and decrypt       & Rotate alphabet for encryption, then reverse the operation.            & Algo.    \\
 Dedup custom objects      & Implement a class with \_\_hash\_\_ and obtain a count unique objects. & class        \\
 Drunken python            & Convert between integer and string without using built-in functions.   & string       \\
 Morse code            & Encode a string into morse code given its conversion rule.         & Algo.    \\
 Two-sum                   & Implement the two-sum problem on a given input pair.                   & Algo.    \\
 k-means                   & Implement and run k-means on sampled points.                           & D.S. \\
 Even odd sum              & Print the sum of even and odd numbers in a list.                       & math         \\
 Shift zeros               & Move all the zeros in a list to the right.                             & array        \\
 Bootstrap 95\% CI     & Calculate the bootstrap 95\% confidence interval of an array.          & D.S. \\
 Sum even digits           & Sum the even digits between two numbers.                               & math         \\
 Min-max diff              & Compute the difference between max and min numbers in a list.          & array        \\
 Distinct chars            & Print the sorted, case-insensitive unique characters of a string.      & string       \\
 Longer string             & Compare and print the longer string given two strings.                 & string       \\
 Sum float digits      & Sum numbers before and after the decimal point of a float.             & math         \\
 Count vowels              & Count the number of vowels in a string.                                & string       \\
 Factorial                 & Compute the factorial of n.                                            & math         \\
 Max edge triangle         & Finds the maximum range of a triangle's third edge.                    & math         \\
 Factorial \& remainder    & Compute the factorial and its remainder when divided.                  & math         \\
 Sum polygon angles        & Sum the angles in a polygon.                                           & math         \\
 Sum string numbers        & Add together two numbers represented in string.                        & string       \\
 Min-max sum               & Sum the range from the minimum to the maximum of a list.               & array        \\
 Vowel overlap             & Find the number of overlapped vowels of two words.                     & string       \\
 Sum negative              & Calculate the sum of negative numbers in a list.                       & math         \\
 Load dataset              & Load from a file and print statistics.                                 & D.S. \\
 Char length list          & Return a list of non-punctuation character lengths from words.         & string       \\
 Hex to RGB                & Convert a six hexadecimal digit string into list of RGB values.        & math         \\
 Majority vote             & Check if a certain element is the majority of a given list.            & array        \\
 Week later                & Print the formatted date of a week later given a date.                 & string       \\
 Sorted word weights       & Check if the list of word weights (sum of ASCII values) are sorted.    & math         \\
 Create Palindrome         & Sum pairs of adjacent digits until the number is palindrome.           & string       \\
 Simulate Backspace    & Apply the backspace characters in a string and print the modified.     & string       \\
 Data manipulation         & Manipulate a pandas DataFrame and split into train and test set.       & D.S. \\
 Sum non-overlap           & Sum the integers in a (min, max) range that don't appear in a list.    & array        \\
 Detect digits             & Find if a string contains digits.                                      & array        \\
 Cascading functions       & Sequentially invoke function objects in a given list.                  & math         \\
 Pluralize duplicates      & Pluralize duplicated words in a list.                                  & dict         \\
 Highest altitude          & Given relative altitudes , find the highest altitude                   & array        \\
 Truncate words            & Truncate a sentence so that it contains k words                        & array        \\
 Single element            & Find the elements that appear one time in an array                     & array        \\
 Remove elements           & Remove all the occurrences of an element in an array                   & array        \\
 Check array sum           & Check whether the sum of an array is equal to a given value            & array        \\
\bottomrule
\end{tabular}
\caption{Problems in \mtpb, showing the problem 1 to 55. D.S. and Algo. refers to data science and algorithm.}
\label{tab:problem_list}
\end{table*}

\begin{table*}[h!]
\footnotesize
\centering
\begin{tabular}{lll} 
\toprule
 Problem Name       & Problem Description                                                                   & Category  \\ \midrule
 Merge sorted lists        & Merge two sorted lists into one                                        & Algo.    \\
 Maximum subarray          & Find the max contiguous subarray and return the sum                    & Algo.    \\
 Max square root integer   & Find the largest integer but smaller than the square root              & Algo.    \\
 Longest word              & Find the longest word in a word list                                   & Algo.    \\
 Sum unique elements       & Sum all the unique numbers in a list                                   & Algo.    \\
 Diagonal sum              & Compute the diagonal sum of a matrix                                   & D.S. \\
 Matrix condition number   & Check condition number of a matrix is less than a threshold            & D.S. \\
 Matrix multiplication sum & Compute matrix multiplication sum of two matrices                      & D.S. \\
 Matrix determinant        & Compare two matrix determinants                                        & D.S. \\
 Log-sum-exp               & Compute the log of sum exponential input                               & D.S. \\
 K nearest points          & Find the k nearest points to the origin                                & array        \\
 Longest common prefix     & Find the longest common prefix of two strings                          & Algo.    \\
 Duplicate elements        & Find duplicates in a list                                              & array        \\
 First unique character    & Find the first non-repeating character in a string                     & Algo.    \\
 Uncommon words            & Find uncommon words in two sentences                                   & Algo.    \\ 
 Average words length              & Compute the average word length of a sentence                                 & Algo.    \\
 Compare char freq                 & Compare the character frequencies in two strings                              & string       \\
 Reverse string                    & Reverse a string                                                              & string       \\
 Square Sum diff                   & Difference between the square of sum and the sum of squares                   & math         \\
 Cosine sim                        & Compute the cosine similarity between two vectors                             & math         \\
 Vector distance                   & Compare vector distances to the origin                                        & math         \\
 Smallest standard dev.            & Find the smaller standard deviation given two lists                           & D.S. \\
 Smallest means                    & Find the smaller mean given two lists                                         & D.S. \\
 Coefficient of variation          & Compute coefficient of variation given a list                                 & D.S. \\
 L1 norm                           & Compute the L1 norm given a list                                              & D.S. \\
 Z-statistic                       & Compute z-statistic given a list                                              & D.S. \\
 Move negatives                    & Move all negative elements in a list to the end                               & array        \\
 Remove alphabets                  & Remove alphabetical characters in a string                                    & string       \\
 Largest norm                      & Find the largest norm among n-dimensional points                              & D.S. \\
 F1 score                          & Given two arrays (pred, gold), calculate the F1 score                         & D.S. \\
 Add Space                         & Add spaces before capital letters                                             & string       \\
 Remove outlier                    & Remove data points in the tail (2sigma) of normal distribution                & D.S. \\
 Convert to categorical            & Convert values into categorical variables                                     & D.S. \\
 Group by key                      & Group items in an array using a provided function                             & array        \\
 Max stock profit                  & Given an array of "prices", find the max profit                               & array        \\
 Sum positions                     & Sum of all position indices where a value appear                              & array        \\
 Find missing num                  & Find a missing number given a list and a max number                           & array        \\
 Common num in matrix              & Common numbers among rows in a matrix                                         & array        \\
 Sum Collatz                       & Obtain the sum of Collatz sequence starting from given number                 & Algo.    \\
 Cup swap                          & Name the location of a "ball" after cup swapping                              & Algo.    \\
 Reverse digits                    & Reverse digits in a number with a stack                                       & Algo.    \\
 Calculate arrows                  & Calculate arrowheads left and right                                           & Algo.    \\
 Check interval num                & Check if the interval (max-min) is included in a list                         & Algo.    \\
 Length encoding                   & Encode a string by converting repeated chars with counts                      & string       \\
 Convert email                     & Use regex to match email addresses and remove special chars                   & string       \\
 Second largest                    & Print out the second largest element in an array                              & array        \\
 Largest prefix sum                & Return the largest prefix sum in an array                                     & array        \\
 Closest element to zero           & Find the element which is the closest to 0 and print the distance             & array        \\
 Consecutive unique char           & Find the max length contiguous subarray with unique characters                & string       \\
 Highest frequency char            & Obtain the frequency of the most frequent character                           & string       \\
 Longest palindrome                & Find the length of longest palindrome substring                               & string       \\
 Count primes                      & Calculate prime numbers in a range                                            & Algo.    \\
 Rotate array                      & Rotate an array to the right k steps                                          & Algo.    \\
 Partition equal sets              & Check if an array can be split into two sets with equal sums                  & Algo.    \\
 Square root integer               & Compute the integer part of square root                                       & math         \\
 Plus 1                            & Return the digits after an integer is added by 1                              & math         \\
 Check square sum                  & Check whether one integer is a sum of  two square numbers                     & math         \\
 Compare standard dev.             & Determine whether standard deviation is less than 1                           & D.S. \\
 Matrix size                       & Calculate the sum of row and column numbers                                   & D.S. \\
 Diff mean and median              & Calculate the difference between mean and median for an array                 & D.S.\\
\bottomrule
\end{tabular}
\caption{Problems in \mtpb, showing the problem 56 to 115. D.S. and Algo. refers to data science and algorithm.}
\label{tab:problem_list_2}
\end{table*}

\clearpage
\section{Perplexity Computation for Single- and Multi-turn Prompts}
\label{app:ppl}
Suppose $\{p_i\}_{i=1}^n$ is the set of prompts for a given problem, and $\{s_i\}_{i=1}^n$ are the $n$ sub-programs synthesized by a model $P_\theta$. Suppose $c_{i-1} = [p_1; s_1; ...; p_{i-1}; s_{i-1}]$ where $[\cdot \  ; \cdot]$ indicates concatenation, the conditional probability of $p_i$ is ${\rm Prob}_i = P_\theta(p_i | c_{i-1})$, and then the perplexity for the multi-turn prompts is computed as 
\begin{align}
    {\rm PPL_{Multi-turn}} = \exp\left(-\frac{1}{m}\sum_{i=1}^n \log {\rm Prob}_i\right),
\end{align}
where $m$ is the total number of tokens of all prompts $\{p_i\}_{i=1}^n$. Suppose $c = [p_1; s_1; ..., p_n, s_n]$, then its probability is ${\rm Prob} = P_\theta (c)$, and the the perplexity for the single-turn prompts is computed as
\begin{align}
    {\rm PPL_{Single-turn}} = \exp\left(-\frac{1}{m}\log {\rm Prob}\right).
\end{align}

\section{Perplexity Comparison for \nl and \pl}
\label{ref:ppl-nl-and-multi}

\begin{table}[H]
\begin{center}
\scalebox{\scale}{
\begin{footnotesize}
\begin{tabular}{p{3cm}ccc}
\toprule
\nl & 350M & 2.7B & 6.1B \\
\midrule
Pass & $4.53$ & $3.25$ & $2.78$ \\
Non-Pass & $4.96$ & $3.87$ & $3.65$ \\
\bottomrule
\end{tabular}
\end{footnotesize}
}
\end{center}
\caption{
Average prompt perplexity$^\downarrow$ of \nl models on pass and non-pass problems.}

\label{tab:humaneval-ppl}
%
\end{table}

\begin{table}[H]
\begin{center}
\scalebox{\scale}{
\begin{footnotesize}
\begin{tabular}{p{3cm}ccc}
\toprule
\pl & 350M & 2.7B & 6.1B \\
\midrule
Pass & $4.78$ & $3.82$ & $3.82$ \\
Non-Pass & $5.64$ & $4.85$ & $4.80$ \\
\bottomrule
\end{tabular}
\end{footnotesize}
}
\end{center}
\caption{
Average prompt perplexity$^\downarrow$ of \pl models on pass and non-pass problems.}

\label{tab:humaneval-ppl}
%
\end{table}

\section{Additional Benchmark Results}
\begin{table}[H]
\begin{center}
\scalebox{\scale}{
\begin{footnotesize}
\begin{tabular}{lccc}
\toprule
Model & pass@$1$ & pass@$10$ & pass@$100$ \\
\midrule
\nlt    & \phantom{0}0.96  & \phantom{0}6.37  & 19.91 \\
\nls      & \phantom{0}5.34  & 24.63 & 48.95 \\
\nll      & \phantom{0}8.15  & 31.21 & 55.27 \\
\nlxl     & 10.92 & 38.43 & 62.76 \\
\plt & \phantom{0}7.46  & 24.18 & 46.37 \\
\pls   & 18.06 & 45.80 & 65.34 \\
\pll   & 18.35 & 47.27 & 67.92 \\
\plxl  & 20.94 & 51.61 & 70.02 \\
\mplt  & 14.59 & 41.49 & 63.00 \\
\mpls    & 27.31 & 59.19 & 74.24 \\
\mpll    & 32.48 & 64.20 & 76.81 \\
\mplxl   & 35.28 & 67.32 & 80.09 \\
\textsc{InCoder 6B}         & 21.30 & 46.50 & 66.20 \\
code-cushman-001   & 45.90 & 66.90 & 79.90 \\
code-davinci-001   & 51.80 & 72.80 & 84.10 \\
code-davinci-002   & 58.10 & 76.70 & 84.50 \\
\bottomrule
\end{tabular}
\end{footnotesize}
}
\end{center}
\caption{
Pass rates on Mostly Basic Python Problems (MBPP).}

\label{tab:mbbp}
\end{table}
We also evaluated our models on Mostly Basic Python Problems (MBPP) \citep{austin2021program}. The results are displayed in Table~\ref{tab:mbbp}. Following \cite{chen2022codet}, we sampled programs from the sanitized MBPP for all of our models, with $n=100$ and temperature$=0.8$. The last four rows are from the aforementioned paper. In general we observe the consistent trend of improving the performance over different versions (NL, Multi, Mono), with our largest \mplxl approaching the results from code-cushman-001. While we do not know whether any of OpenAI models is the “Codex 12B” reported in \cite{chen2021evaluating}, we believe our model achieves reasonable results on MBPP as well. We also note that our \mpll significantly outperformed \textsc{InCoder 6B}.

\section{Generated Samples}
\label{app:qualitative_examples}

\subsection{Cases where \mplxl under-performs}

\input{code_codegen/prob96}
\input{code_codegen/prob62}

\clearpage
\subsection{Cases where \mplxl outperforms}

\input{code_codegen/prob95}
\input{code_codegen/prob48}

\newpage
\section{Additional Analyses on MTPB}

\begin{figure*}[t]
	\centering	
	\includegraphics[width=0.75\linewidth]{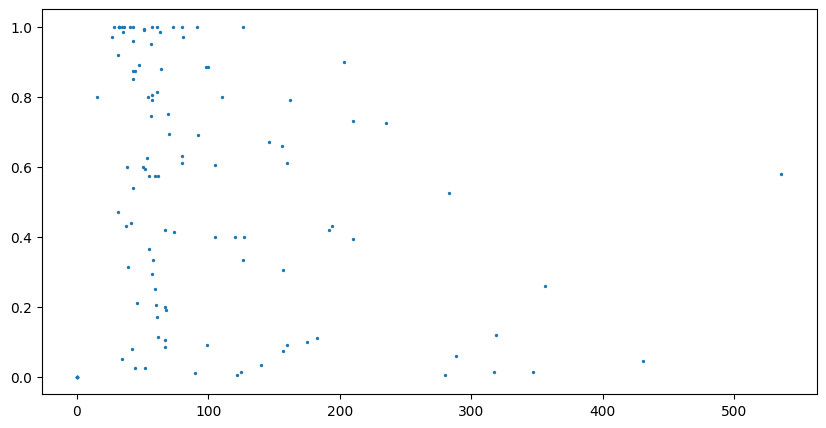}
	\caption{Maximum Length of Completion versus Pass Rate.}
	\label{fig:length-pass-rate-1}
\end{figure*}

\begin{figure*}[t]
	\centering	
	\includegraphics[width=0.75\linewidth]{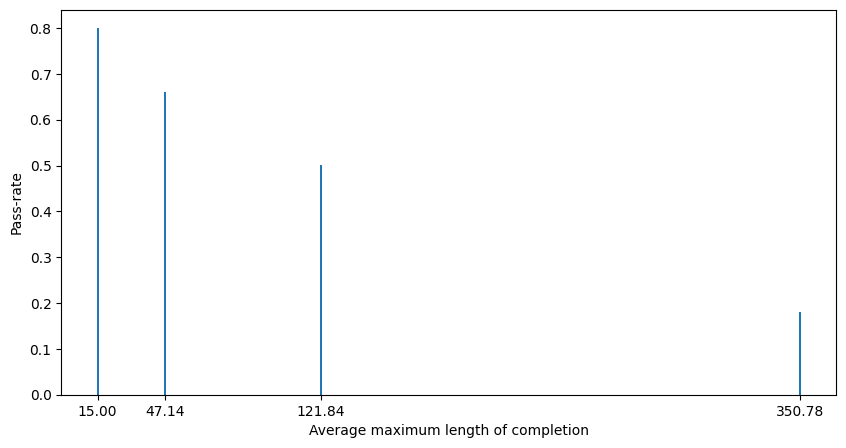}
	\caption{Maximum Length of Completion versus Pass Rate.}
	\label{fig:length-pass-rate-2}
\end{figure*}

\begin{figure*}[t]
	\centering	
	\includegraphics[width=0.75\linewidth]{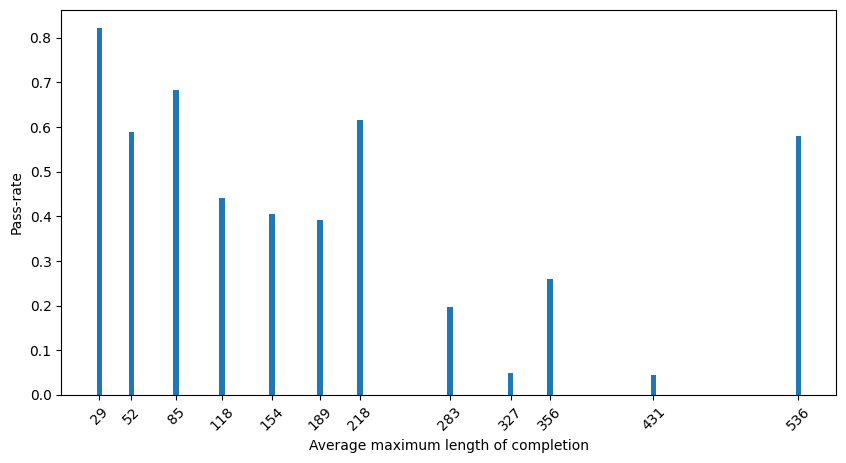}
	\caption{Maximum Length of Completion versus Pass Rate.}
	\label{fig:length-pass-rate-3}
\end{figure*}

\begin{figure*}[t]
	\centering	
	\includegraphics[width=0.75\linewidth]{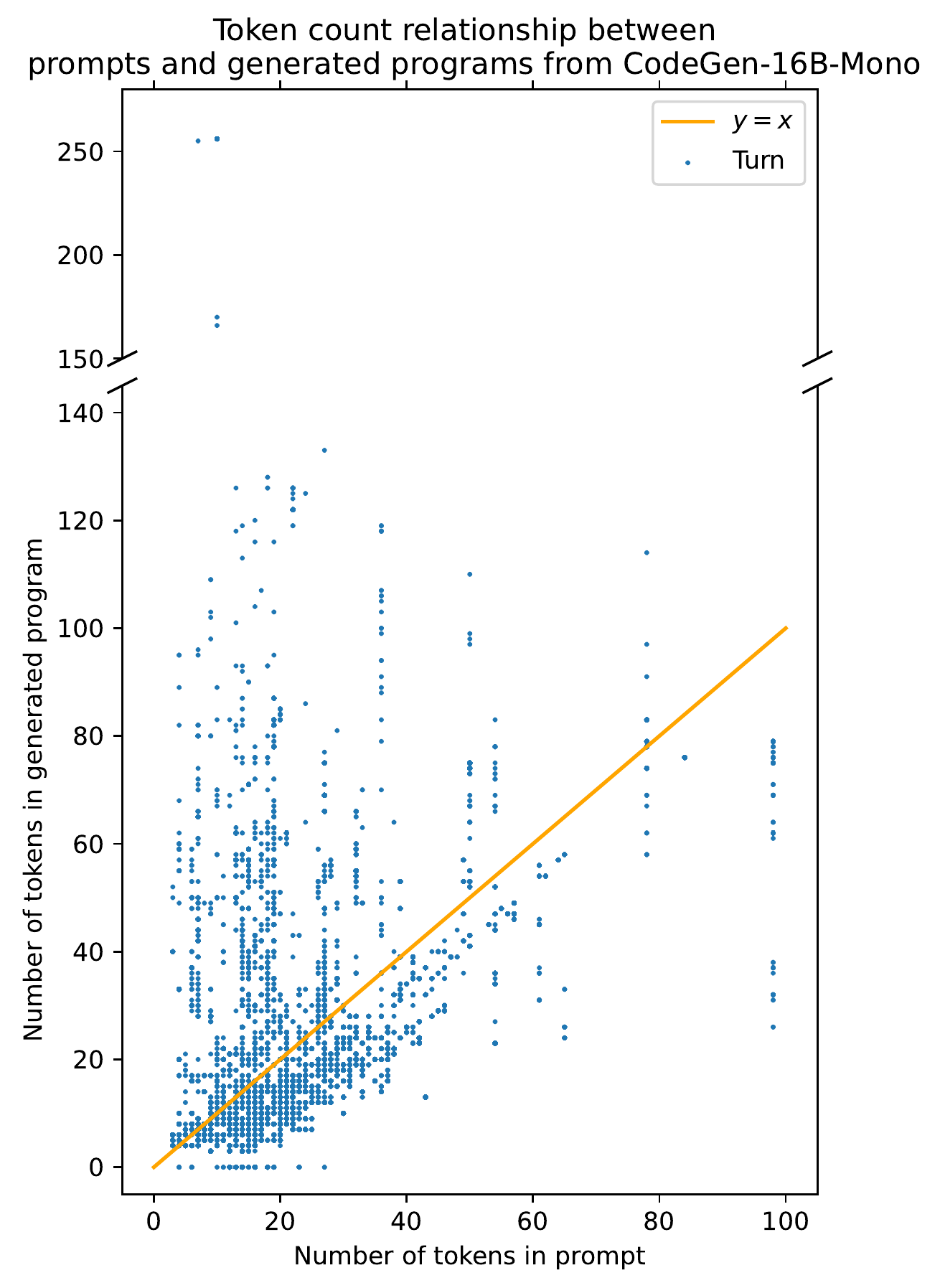}
	\caption{Prompt Length versus Generated Program Length.}
	\label{fig:code_len_vs_prompt_len}
\end{figure*}

We conducted additional analyses to illustrate the relationship generated program length and pass rate and showed the results in Figure~\ref{fig:length-pass-rate-1}, Figure~\ref{fig:length-pass-rate-2}, and Figure~\ref{fig:length-pass-rate-3}. The relationship between generated program length and prompt length is shown in Figure~\ref{fig:code_len_vs_prompt_len}.

\end{document}

%% file: math_commands.tex
\usepackage{amsmath,amsfonts,bm}

\def\eqref#1{equation~\ref{#1}}

\def\1{\bm{1}}

\DeclareMathAlphabet{\mathsfit}{\encodingdefault}{\sfdefault}{m}{sl}
\SetMathAlphabet{\mathsfit}{bold}{\encodingdefault}{\sfdefault}{bx}{n}



%% file: code_codegen/prob96.tex
\begin{figure*}[h]
\centering
\setlength{\tabcolsep}{5pt}
\begin{tabular}{@{}p{0.49\linewidth}p{0.49\linewidth}@{}}
\begin{lstlisting}[language=Python,linebackgroundcolor={%
    \btLstHL{8}{\color{ll1}}%
}]
# Import libraries.

import numpy as np

# Initialize a variable "stack" with an empty list, and "num" with 100 as a string.

stack = []
num = "100"

# For each chracter in "num", append the character to "stack".

for i in range(len(num)):
    stack.append(num[i])

# Assign an empty string to a variable "result", and concatenate characters popped from the last element of "stack" to "result" until "stack" is empty.

result = ""
while len(stack)!= 0:
    result += stack.pop()

# Cast "result" as integer and print it out.

print(int(result))
  \end{lstlisting}&
  \begin{lstlisting}[language=Python,linebackgroundcolor={%
    \btLstHL{8}{\color{ll2}}%
}]
# Import libraries.

import numpy as np

# Initialize a variable "stack" with an empty list, and "num" with 100 as a string.

stack = []
num = 100

# For each chracter in "num", append the character to "stack".

for i in num:
    stack.append(i)

# Assign an empty string to a variable "result", and concatenate characters popped from the last element of "stack" to "result" until "stack" is empty.

result = ""
while len(stack) > 0:
    result += stack.pop()

# Cast "result" as integer and print it out.

print(int(result))
  \end{lstlisting}\\
  \caption*{\mpls} & \caption*{\mplxl}
 \end{tabular}
 \caption{Generated samples for ``Reverse digits'' problem. While being simplistic, we often observe that \mplxl assumes the type when assigning a number to a variable. Here on the right, despite being explicitly specified to initialize a number as a string, the larger model fails to do so. This causes an error in the following turn when the code attempts to iterate over \texttt{num}.}
  \label{lst:problem_96}
\end{figure*}

%% file: code_codegen/prob62.tex
\begin{figure*}[h]
\centering
\setlength{\tabcolsep}{5pt}
\begin{tabular}{@{}p{0.49\linewidth}p{0.49\linewidth}@{}}
\begin{lstlisting}[language=Python,linebackgroundcolor={%
    \btLstHL{19-22}{\color{ll1}}%
}]
# Import libraries.

import numpy as np

# Assign the matrix "[[3, 2], [2, 3]]" to a variable named "my_matrix".

my_matrix = [[3, 2], [2, 3]]

# Assign the number "1" to a variable named "t".

t = 1

# Compute the condition number of my_matrix and store as result.

result = np.linalg.cond(my_matrix)

# Check whether the result is smaller than t. If yes, return "True", otherwise return "False".

if result < t:
    print("True")
else:
    print("False")
  \end{lstlisting}&
  \begin{lstlisting}[language=Python,linebackgroundcolor={%
    \btLstHL{19-23}{\color{ll2}}%
}]
# Import libraries.

import numpy as np

# Assign the matrix "[[3, 2], [2, 3]]" to a variable named "my_matrix".

my_matrix = [[3, 2], [2, 3]]

# Assign the number "1" to a variable named "t".

t = 1

# Compute the condition number of my_matrix and store as result.

condition_number = np.linalg.cond(my_matrix)

# Check whether the result is smaller than t. If yes, return "True", otherwise return "False".

def check_condition_number(condition_number, t):
    if condition_number < t:
        return True
    else:
        return False
  \end{lstlisting}\\
  \caption*{\mpls} & \caption*{\mplxl}
 \end{tabular}
 \caption{Generated samples for ``Matrix condition number'' problem. Both models generate programs in the same scope until the last turn, where \mplxl generates in the form of function definition. Noticeably, the larger model generates \texttt{return} keyword that appears in the prompt, but it is only valid within a function.}
  \label{lst:problem_62}
\end{figure*}

%% file: code_codegen/prob95.tex
\begin{figure*}[h]
\centering
\setlength{\tabcolsep}{5pt}
\begin{tabular}{@{}p{0.49\linewidth}p{0.49\linewidth}@{}}
\begin{lstlisting}[language=Python,linebackgroundcolor={%
    \btLstHL{7}{\color{ll2}}%
}]
# Import libraries.

import numpy as np

# Define a variable "pos" with "B", "swap" with ['AC', 'CA'].

pos = ['B','swap']

# Write a function "move" that takes two strings x and y as input, and replace any appearance of x in y with an empty string, then return y.

def move(x,y):
    return y.replace(x,'')

# For each element in "swap", if it contains "pos", call "move" on "pos" and the current element and store the result to "pos".

def swap(swap):
    for i in range(len(swap)):
        if swap[i] == 'pos':
            swap[i] = move(pos[0],pos[1])
    return swap

# Print out "pos".

print(pos)
  \end{lstlisting}&
  \begin{lstlisting}[language=Python,linebackgroundcolor={%
    \btLstHL{7-8}{\color{ll1}}%
}]
# Import libraries.

import numpy as np

# Define a variable "pos" with "B", "swap" with ['AC', 'CA'].

pos = "B"
swap = ['AC', 'CA']

# Write a function "move" that takes two strings x and y as input, and replace any appearance of x in y with an empty string, then return y.

def move(x, y):
    return y.replace(x, "")

# For each element in "swap", if it contains "pos", call "move" on "pos" and the current element and store the result to "pos".

for i in swap:
    if pos in i:
        pos = move(pos, i)

# Print out "pos".

print(pos)
  \end{lstlisting}\\
  \caption*{\mpls} & \caption*{\mplxl}
 \end{tabular}
 \caption{Generated samples for ``Cup swap'' problem. Here, \mpls mistakenly treats ``swap'' as a part of a value t o be assigned, while \mplxl parses the intent correctly and assigns two variables in parallel.}
  \label{lst:problem_95}
\end{figure*}

%% file: code_codegen/prob48.tex
\begin{figure*}[h]
\centering
\setlength{\tabcolsep}{5pt}
\begin{tabular}{@{}p{0.49\linewidth}p{0.49\linewidth}@{}}
\begin{lstlisting}[language=Python,linebackgroundcolor={%
    \btLstHL{11-16}{\color{ll2}}%
}]
# Import libraries.

import numpy as np

# Initialize the variable named lst1 with a list ['abc', 'abc10'].

lst1 = ['abc', 'abc10']

# Create a function called num_in_str() to check whether a string contains a number.

def num_in_str(string):
    for i in string:
        if i.isnumeric():
            return True
        else:
            return False

# Call the function num_in_str() to find strings in lst1 that have numbers and assign them to a list named lst2

lst2 = []

for i in lst1:
    if num_in_str(i):
        lst2.append(i)

# Print out lst2

print(lst2)
  \end{lstlisting}&
  \begin{lstlisting}[language=Python,linebackgroundcolor={%
    \btLstHL{11-12}{\color{ll1}}%
}]
# Import libraries.

import numpy as np

# Initialize the variable named lst1 with a list ['abc', 'abc10'].

lst1 = ['abc', 'abc10']

# Create a function called num_in_str() to check whether a string contains a number.

def num_in_str(s):
    return any(i.isdigit() for i in s)

# Call the function num_in_str() to find strings in lst1 that have numbers and assign them to a list named lst2

lst2 = [i for i in lst1 if num_in_str(i)]

# Print out lst2

print(lst2)
  \end{lstlisting}\\
  \caption*{\mpls} & \caption*{\mplxl}
 \end{tabular}
 \caption{Generated samples for ``Detect digits'' problem. \mpls fails to implement the \texttt{num\_in\_str}, resulting in checking only the first character. In contrast, \mplxl successfully utilizes \texttt{any} function to scan all the characters in the given string.}
  \label{lst:problem_48}
\end{figure*}

%% file: iclr2023_conference.bbl
\begin{thebibliography}{48}
\providecommand{\natexlab}[1]{#1}
\providecommand{\url}[1]{\texttt{#1}}
\expandafter\ifx\csname urlstyle\endcsname\relax
  \providecommand{\doi}[1]{doi: #1}\else
  \providecommand{\doi}{doi: \begingroup \urlstyle{rm}\Url}\fi

\bibitem[Agashe et~al.(2019)Agashe, Iyer, and Zettlemoyer]{agashe2019juice}
Rajas Agashe, Srinivasan Iyer, and Luke Zettlemoyer.
\newblock Juice: A large scale distantly supervised dataset for open domain
  context-based code generation.
\newblock In \emph{Proceedings of the 2019 Conference on Empirical Methods in
  Natural Language Processing and the 9th International Joint Conference on
  Natural Language Processing (EMNLP-IJCNLP)}, pp.\  5436--5446, 2019.

\bibitem[Andreas et~al.(2020)Andreas, Bufe, Burkett, Chen, Clausman, Crawford,
  Crim, DeLoach, Dorner, Eisner, et~al.]{andreas2020task}
Jacob Andreas, John Bufe, David Burkett, Charles Chen, Josh Clausman, Jean
  Crawford, Kate Crim, Jordan DeLoach, Leah Dorner, Jason Eisner, et~al.
\newblock Task-oriented dialogue as dataflow synthesis.
\newblock \emph{Transactions of the Association for Computational Linguistics},
  8:\penalty0 556--571, 2020.

\bibitem[Austin et~al.(2021)Austin, Odena, Nye, Bosma, Michalewski, Dohan,
  Jiang, Cai, Terry, Le, et~al.]{austin2021program}
Jacob Austin, Augustus Odena, Maxwell Nye, Maarten Bosma, Henryk Michalewski,
  David Dohan, Ellen Jiang, Carrie Cai, Michael Terry, Quoc Le, et~al.
\newblock Program synthesis with large language models.
\newblock \emph{arXiv preprint arXiv:2108.07732}, 2021.

\bibitem[Bahdanau et~al.(2014)Bahdanau, Cho, and Bengio]{bahdanau2014neural}
Dzmitry Bahdanau, Kyunghyun Cho, and Yoshua Bengio.
\newblock Neural machine translation by jointly learning to align and
  translate.
\newblock \emph{arXiv preprint arXiv:1409.0473}, 2014.

\bibitem[Black et~al.(2021)Black, Gao, Wang, Leahy, and Biderman]{gpt-neo}
Sid Black, Leo Gao, Phil Wang, Connor Leahy, and Stella Biderman.
\newblock {GPT-Neo: Large Scale Autoregressive Language Modeling with
  Mesh-Tensorflow}, March 2021.
\newblock URL \url{https://doi.org/10.5281/zenodo.5297715}.
\newblock {If you use this software, please cite it using these metadata.}

\bibitem[Bradbury et~al.(2018)Bradbury, Frostig, Hawkins, Johnson, Leary,
  Maclaurin, Necula, Paszke, Vander{P}las, Wanderman-{M}ilne, and
  Zhang]{jax2018github}
James Bradbury, Roy Frostig, Peter Hawkins, Matthew~James Johnson, Chris Leary,
  Dougal Maclaurin, George Necula, Adam Paszke, Jake Vander{P}las, Skye
  Wanderman-{M}ilne, and Qiao Zhang.
\newblock {JAX}: composable transformations of {P}ython+{N}um{P}y programs,
  2018.
\newblock URL \url{http://github.com/google/jax}.

\bibitem[Brown et~al.(2020)Brown, Mann, Ryder, Subbiah, Kaplan, Dhariwal,
  Neelakantan, Shyam, Sastry, Askell, et~al.]{brown2020language}
Tom Brown, Benjamin Mann, Nick Ryder, Melanie Subbiah, Jared~D Kaplan, Prafulla
  Dhariwal, Arvind Neelakantan, Pranav Shyam, Girish Sastry, Amanda Askell,
  et~al.
\newblock Language models are few-shot learners.
\newblock \emph{Advances in neural information processing systems},
  33:\penalty0 1877--1901, 2020.

\bibitem[Chen et~al.(2022)Chen, Zhang, Nguyen, Zan, Lin, Lou, and
  Chen]{chen2022codet}
Bei Chen, Fengji Zhang, Anh Nguyen, Daoguang Zan, Zeqi Lin, Jian-Guang Lou, and
  Weizhu Chen.
\newblock Codet: Code generation with generated tests.
\newblock \emph{arXiv preprint arXiv:2207.10397}, 2022.

\bibitem[Chen et~al.(2021)Chen, Tworek, Jun, Yuan, Pinto, Kaplan, Edwards,
  Burda, Joseph, Brockman, et~al.]{chen2021evaluating}
Mark Chen, Jerry Tworek, Heewoo Jun, Qiming Yuan, Henrique Ponde de~Oliveira
  Pinto, Jared Kaplan, Harri Edwards, Yuri Burda, Nicholas Joseph, Greg
  Brockman, et~al.
\newblock Evaluating large language models trained on code.
\newblock \emph{arXiv preprint arXiv:2107.03374}, 2021.

\bibitem[Cheung et~al.(2013)Cheung, Solar-Lezama, and
  Madden]{cheung2013optimizing}
Alvin Cheung, Armando Solar-Lezama, and Samuel Madden.
\newblock Optimizing database-backed applications with query synthesis.
\newblock \emph{ACM SIGPLAN Notices}, 48\penalty0 (6):\penalty0 3--14, 2013.

\bibitem[Clement et~al.(2020)Clement, Drain, Timcheck, Svyatkovskiy, and
  Sundaresan]{clement2020pymt5}
Colin Clement, Dawn Drain, Jonathan Timcheck, Alexey Svyatkovskiy, and Neel
  Sundaresan.
\newblock Pymt5: multi-mode translation of natural language and python code
  with transformers.
\newblock In \emph{Proceedings of the 2020 Conference on Empirical Methods in
  Natural Language Processing (EMNLP)}, pp.\  9052--9065, 2020.

\bibitem[Devlin et~al.(2019)Devlin, Chang, Lee, and Toutanova]{devlin2018bert}
Jacob Devlin, Ming-Wei Chang, Kenton Lee, and Kristina Toutanova.
\newblock {BERT}: Pre-training of deep bidirectional transformers for language
  understanding.
\newblock In \emph{Proceedings of the 2019 Conference of the North {A}merican
  Chapter of the Association for Computational Linguistics: Human Language
  Technologies, Volume 1 (Long and Short Papers)}, pp.\  4171--4186,
  Minneapolis, Minnesota, June 2019. Association for Computational Linguistics.
\newblock \doi{10.18653/v1/N19-1423}.
\newblock URL \url{https://aclanthology.org/N19-1423}.

\bibitem[Dosovitskiy et~al.(2021)Dosovitskiy, Beyer, Kolesnikov, Weissenborn,
  Zhai, Unterthiner, Dehghani, Minderer, Heigold, Gelly, Uszkoreit, and
  Houlsby]{dosovitskiy2020image}
Alexey Dosovitskiy, Lucas Beyer, Alexander Kolesnikov, Dirk Weissenborn,
  Xiaohua Zhai, Thomas Unterthiner, Mostafa Dehghani, Matthias Minderer, Georg
  Heigold, Sylvain Gelly, Jakob Uszkoreit, and Neil Houlsby.
\newblock An image is worth 16x16 words: Transformers for image recognition at
  scale.
\newblock In \emph{ICLR}, 2021.
\newblock URL \url{https://openreview.net/forum?id=YicbFdNTTy}.

\bibitem[Feng et~al.(2020)Feng, Guo, Tang, Duan, Feng, Gong, Shou, Qin, Liu,
  Jiang, et~al.]{feng2020codebert}
Zhangyin Feng, Daya Guo, Duyu Tang, Nan Duan, Xiaocheng Feng, Ming Gong, Linjun
  Shou, Bing Qin, Ting Liu, Daxin Jiang, et~al.
\newblock Codebert: A pre-trained model for programming and natural languages.
\newblock In \emph{Findings of the Association for Computational Linguistics:
  EMNLP 2020}, pp.\  1536--1547, 2020.

\bibitem[Fried et~al.(2022)Fried, Aghajanyan, Lin, Wang, Wallace, Shi, Zhong,
  Yih, Zettlemoyer, and Lewis]{fried2022incoder}
Daniel Fried, Armen Aghajanyan, Jessy Lin, Sida Wang, Eric Wallace, Freda Shi,
  Ruiqi Zhong, Wen-tau Yih, Luke Zettlemoyer, and Mike Lewis.
\newblock Incoder: A generative model for code infilling and synthesis.
\newblock \emph{arXiv preprint arXiv:2204.05999}, 2022.

\bibitem[Gao et~al.(2020)Gao, Biderman, Black, Golding, Hoppe, Foster, Phang,
  He, Thite, Nabeshima, et~al.]{gao2020pile}
Leo Gao, Stella Biderman, Sid Black, Laurence Golding, Travis Hoppe, Charles
  Foster, Jason Phang, Horace He, Anish Thite, Noa Nabeshima, et~al.
\newblock The pile: An 800gb dataset of diverse text for language modeling.
\newblock \emph{arXiv preprint arXiv:2101.00027}, 2020.

\bibitem[Gulwani(2011)]{gulwani2011automating}
Sumit Gulwani.
\newblock Automating string processing in spreadsheets using input-output
  examples.
\newblock \emph{ACM Sigplan Notices}, 46\penalty0 (1):\penalty0 317--330, 2011.

\bibitem[Gulwani et~al.(2017)Gulwani, Polozov, Singh,
  et~al.]{gulwani2017program}
Sumit Gulwani, Oleksandr Polozov, Rishabh Singh, et~al.
\newblock Program synthesis.
\newblock \emph{Foundations and Trends{\textregistered} in Programming
  Languages}, 4\penalty0 (1-2):\penalty0 1--119, 2017.

\bibitem[Hendrycks et~al.(2021)Hendrycks, Basart, Kadavath, Mazeika, Arora,
  Guo, Burns, Puranik, He, Song, and Steinhardt]{hendrycks2021measuring}
Dan Hendrycks, Steven Basart, Saurav Kadavath, Mantas Mazeika, Akul Arora,
  Ethan Guo, Collin Burns, Samir Puranik, Horace He, Dawn Song, and Jacob
  Steinhardt.
\newblock Measuring coding challenge competence with {APPS}.
\newblock In \emph{Thirty-fifth Conference on Neural Information Processing
  Systems Datasets and Benchmarks Track (Round 2)}, 2021.
\newblock URL \url{https://openreview.net/forum?id=sD93GOzH3i5}.

\bibitem[Holtzman et~al.(2020)Holtzman, Buys, Du, Forbes, and
  Choi]{holtzman2019curious}
Ari Holtzman, Jan Buys, Li~Du, Maxwell Forbes, and Yejin Choi.
\newblock The curious case of neural text degeneration.
\newblock In \emph{ICLR}, 2020.
\newblock URL \url{https://openreview.net/forum?id=rygGQyrFvH}.

\bibitem[Iyer et~al.(2018)Iyer, Konstas, Cheung, and
  Zettlemoyer]{iyer-etal-2018-mapping}
Srinivasan Iyer, Ioannis Konstas, Alvin Cheung, and Luke Zettlemoyer.
\newblock Mapping language to code in programmatic context.
\newblock In \emph{Proceedings of the 2018 Conference on Empirical Methods in
  Natural Language Processing}, pp.\  1643--1652, Brussels, Belgium,
  October-November 2018. Association for Computational Linguistics.
\newblock \doi{10.18653/v1/D18-1192}.
\newblock URL \url{https://aclanthology.org/D18-1192}.

\bibitem[Joshi et~al.(2002)Joshi, Nelson, and Randall]{joshi2002denali}
Rajeev Joshi, Greg Nelson, and Keith Randall.
\newblock Denali: A goal-directed superoptimizer.
\newblock \emph{ACM SIGPLAN Notices}, 37\penalty0 (5):\penalty0 304--314, 2002.

\bibitem[Jumper et~al.(2021)Jumper, Evans, Pritzel, Green, Figurnov,
  Ronneberger, Tunyasuvunakool, Bates, {\v{Z}}{\'\i}dek, Potapenko,
  et~al.]{jumper2021highly}
John Jumper, Richard Evans, Alexander Pritzel, Tim Green, Michael Figurnov,
  Olaf Ronneberger, Kathryn Tunyasuvunakool, Russ Bates, Augustin
  {\v{Z}}{\'\i}dek, Anna Potapenko, et~al.
\newblock Highly accurate protein structure prediction with alphafold.
\newblock \emph{Nature}, 596\penalty0 (7873):\penalty0 583--589, 2021.

\bibitem[Kanade et~al.(2020)Kanade, Maniatis, Balakrishnan, and
  Shi]{kanade2020learning}
Aditya Kanade, Petros Maniatis, Gogul Balakrishnan, and Kensen Shi.
\newblock Learning and evaluating contextual embedding of source code.
\newblock In \emph{International Conference on Machine Learning}, pp.\
  5110--5121. PMLR, 2020.

\bibitem[Kingma \& Ba(2015)Kingma and Ba]{kingma2014adam}
Diederik~P. Kingma and Jimmy Ba.
\newblock Adam: A method for stochastic optimization.
\newblock In \emph{ICLR (Poster)}, 2015.
\newblock URL \url{http://arxiv.org/abs/1412.6980}.

\bibitem[Kulal et~al.(2019)Kulal, Pasupat, Chandra, Lee, Padon, Aiken, and
  Liang]{kulal2019spoc}
Sumith Kulal, Panupong Pasupat, Kartik Chandra, Mina Lee, Oded Padon, Alex
  Aiken, and Percy~S Liang.
\newblock Spoc: Search-based pseudocode to code.
\newblock \emph{Advances in Neural Information Processing Systems}, 32, 2019.

\bibitem[Lewis et~al.(2020)Lewis, Liu, Goyal, Ghazvininejad, Mohamed, Levy,
  Stoyanov, and Zettlemoyer]{lewis2020bart}
Mike Lewis, Yinhan Liu, Naman Goyal, Marjan Ghazvininejad, Abdelrahman Mohamed,
  Omer Levy, Veselin Stoyanov, and Luke Zettlemoyer.
\newblock Bart: Denoising sequence-to-sequence pre-training for natural
  language generation, translation, and comprehension.
\newblock In \emph{Proceedings of the 58th Annual Meeting of the Association
  for Computational Linguistics}, pp.\  7871--7880, 2020.

\bibitem[Li et~al.(2022)Li, Choi, Chung, Kushman, Schrittwieser, Leblond,
  Eccles, Keeling, Gimeno, Dal~Lago, Hubert, Choy, de~Masson~d'Autume,
  Babuschkin, Chen, Huang, Welbl, Gowal, Cherepanov, Molloy, Mankowitz,
  Sutherland~Robson, Kohli, de~Freitas, Kavukcuoglu, and Vinyals]{alphacode}
Yujia Li, David Choi, Junyoung Chung, Nate Kushman, Julian Schrittwieser, Rémi
  Leblond, Tom Eccles, James Keeling, Felix Gimeno, Agustin Dal~Lago, Thomas
  Hubert, Peter Choy, Cyprien de~Masson~d'Autume, Igor Babuschkin, Xinyun Chen,
  Po-Sen Huang, Johannes Welbl, Sven Gowal, Alexey Cherepanov, James Molloy,
  Daniel Mankowitz, Esme Sutherland~Robson, Pushmeet Kohli, Nando de~Freitas,
  Koray Kavukcuoglu, and Oriol Vinyals.
\newblock Competition-level code generation with alphacode, Feb 2022.

\bibitem[Manna \& Waldinger(1971)Manna and Waldinger]{manna1971toward}
Zohar Manna and Richard~J Waldinger.
\newblock Toward automatic program synthesis.
\newblock \emph{Communications of the ACM}, 14\penalty0 (3):\penalty0 151--165,
  1971.

\bibitem[Nye et~al.(2021)Nye, Andreassen, Gur-Ari, Michalewski, Austin, Bieber,
  Dohan, Lewkowycz, Bosma, Luan, et~al.]{nye2021show}
Maxwell Nye, Anders~Johan Andreassen, Guy Gur-Ari, Henryk Michalewski, Jacob
  Austin, David Bieber, David Dohan, Aitor Lewkowycz, Maarten Bosma, David
  Luan, et~al.
\newblock Show your work: Scratchpads for intermediate computation with
  language models.
\newblock \emph{arXiv preprint arXiv:2112.00114}, 2021.

\bibitem[Oord et~al.(2018)Oord, Li, and Vinyals]{oord2018representation}
Aaron van~den Oord, Yazhe Li, and Oriol Vinyals.
\newblock Representation learning with contrastive predictive coding.
\newblock \emph{arXiv preprint arXiv:1807.03748}, 2018.

\bibitem[Panchekha et~al.(2015)Panchekha, Sanchez-Stern, Wilcox, and
  Tatlock]{panchekha2015automatically}
Pavel Panchekha, Alex Sanchez-Stern, James~R Wilcox, and Zachary Tatlock.
\newblock Automatically improving accuracy for floating point expressions.
\newblock \emph{ACM SIGPLAN Notices}, 50\penalty0 (6):\penalty0 1--11, 2015.

\bibitem[Parisotto et~al.(2017)Parisotto, rahman Mohamed, Singh, Li, Zhou, and
  Kohli]{parisotto2016neuro}
Emilio Parisotto, Abdel rahman Mohamed, Rishabh Singh, Lihong Li, Dengyong
  Zhou, and Pushmeet Kohli.
\newblock Neuro-symbolic program synthesis.
\newblock In \emph{ICLR (Poster)}, 2017.
\newblock URL \url{https://openreview.net/forum?id=rJ0JwFcex}.

\bibitem[Pascanu et~al.(2013)Pascanu, Mikolov, and
  Bengio]{pascanu2013difficulty}
Razvan Pascanu, Tomas Mikolov, and Yoshua Bengio.
\newblock On the difficulty of training recurrent neural networks.
\newblock In \emph{International conference on machine learning}, pp.\
  1310--1318. PMLR, 2013.

\bibitem[Polozov \& Gulwani(2015)Polozov and Gulwani]{polozov2015flashmeta}
Oleksandr Polozov and Sumit Gulwani.
\newblock Flashmeta: A framework for inductive program synthesis.
\newblock In \emph{Proceedings of the 2015 ACM SIGPLAN International Conference
  on Object-Oriented Programming, Systems, Languages, and Applications}, pp.\
  107--126, 2015.

\bibitem[Raffel et~al.(2020)Raffel, Shazeer, Roberts, Lee, Narang, Matena,
  Zhou, Li, and Liu]{raffel2020exploring}
Colin Raffel, Noam Shazeer, Adam Roberts, Katherine Lee, Sharan Narang, Michael
  Matena, Yanqi Zhou, Wei Li, and Peter~J Liu.
\newblock Exploring the limits of transfer learning with a unified text-to-text
  transformer.
\newblock \emph{Journal of Machine Learning Research}, 21:\penalty0 1--67,
  2020.

\bibitem[Rajbhandari et~al.(2020)Rajbhandari, Rasley, Ruwase, and
  He]{rajbhandari2020zero}
Samyam Rajbhandari, Jeff Rasley, Olatunji Ruwase, and Yuxiong He.
\newblock Zero: Memory optimizations toward training trillion parameter models.
\newblock In \emph{SC20: International Conference for High Performance
  Computing, Networking, Storage and Analysis}, pp.\  1--16. IEEE, 2020.

\bibitem[Raychev et~al.(2016)Raychev, Bielik, and
  Vechev]{raychev2016probabilistic}
Veselin Raychev, Pavol Bielik, and Martin Vechev.
\newblock Probabilistic model for code with decision trees.
\newblock \emph{ACM SIGPLAN Notices}, 51\penalty0 (10):\penalty0 731--747,
  2016.

\bibitem[Schkufza et~al.(2013)Schkufza, Sharma, and
  Aiken]{schkufza2013stochastic}
Eric Schkufza, Rahul Sharma, and Alex Aiken.
\newblock Stochastic superoptimization.
\newblock \emph{ACM SIGARCH Computer Architecture News}, 41\penalty0
  (1):\penalty0 305--316, 2013.

\bibitem[Shoeybi et~al.(2019)Shoeybi, Patwary, Puri, LeGresley, Casper, and
  Catanzaro]{shoeybi2019megatron}
Mohammad Shoeybi, Mostofa Patwary, Raul Puri, Patrick LeGresley, Jared Casper,
  and Bryan Catanzaro.
\newblock Megatron-lm: Training multi-billion parameter language models using
  model parallelism.
\newblock \emph{arXiv preprint arXiv:1909.08053}, 2019.

\bibitem[Su et~al.(2021)Su, Lu, Pan, Wen, and Liu]{su2021roformer}
Jianlin Su, Yu~Lu, Shengfeng Pan, Bo~Wen, and Yunfeng Liu.
\newblock Roformer: Enhanced transformer with rotary position embedding.
\newblock \emph{arXiv preprint arXiv:2104.09864}, 2021.

\bibitem[Vaithilingam et~al.(2022)Vaithilingam, Zhang, and
  Glassman]{vaithilingam2022expectation}
Priyan Vaithilingam, Tianyi Zhang, and Elena~L Glassman.
\newblock Expectation vs. experience: Evaluating the usability of code
  generation tools powered by large language models.
\newblock In \emph{CHI Conference on Human Factors in Computing Systems
  Extended Abstracts}, pp.\  1--7, 2022.

\bibitem[Vaswani et~al.(2017)Vaswani, Shazeer, Parmar, Uszkoreit, Jones, Gomez,
  Kaiser, and Polosukhin]{vaswani2017attention}
Ashish Vaswani, Noam Shazeer, Niki Parmar, Jakob Uszkoreit, Llion Jones,
  Aidan~N Gomez, {\L}ukasz Kaiser, and Illia Polosukhin.
\newblock Attention is all you need.
\newblock In \emph{Advances in neural information processing systems}, pp.\
  5998--6008, 2017.

\bibitem[Wang \& Komatsuzaki(2021)Wang and Komatsuzaki]{gpt-j}
Ben Wang and Aran Komatsuzaki.
\newblock {GPT-J-6B: A 6 Billion Parameter Autoregressive Language Model}.
\newblock \url{https://github.com/kingoflolz/mesh-transformer-jax}, May 2021.

\bibitem[Wang et~al.(2021)Wang, Wang, Joty, and Hoi]{wang2021codet5}
Yue Wang, Weishi Wang, Shafiq Joty, and Steven~C.H. Hoi.
\newblock Codet5: Identifier-aware unified pre-trained encoder-decoder models
  for code understanding and generation.
\newblock In \emph{Proceedings of the 2021 Conference on Empirical Methods in
  Natural Language Processing, EMNLP 2021}, 2021.

\bibitem[Wei et~al.(2022)Wei, Wang, Schuurmans, Bosma, Chi, Le, and
  Zhou]{wei2022chain}
Jason Wei, Xuezhi Wang, Dale Schuurmans, Maarten Bosma, Ed~Chi, Quoc Le, and
  Denny Zhou.
\newblock Chain of thought prompting elicits reasoning in large language
  models.
\newblock \emph{arXiv preprint arXiv:2201.11903}, 2022.

\bibitem[Yu et~al.(2019{\natexlab{a}})Yu, Zhang, Er, Li, Xue, Pang, Lin, Tan,
  Shi, Li, Jiang, Yasunaga, Shim, Chen, Fabbri, Li, Chen, Zhang, Dixit, Zhang,
  Xiong, Socher, Lasecki, and Radev]{yu-etal-2019-cosql}
Tao Yu, Rui Zhang, Heyang Er, Suyi Li, Eric Xue, Bo~Pang, Xi~Victoria Lin,
  Yi~Chern Tan, Tianze Shi, Zihan Li, Youxuan Jiang, Michihiro Yasunaga,
  Sungrok Shim, Tao Chen, Alexander Fabbri, Zifan Li, Luyao Chen, Yuwen Zhang,
  Shreya Dixit, Vincent Zhang, Caiming Xiong, Richard Socher, Walter Lasecki,
  and Dragomir Radev.
\newblock {C}o{SQL}: A conversational text-to-{SQL} challenge towards
  cross-domain natural language interfaces to databases.
\newblock In \emph{Proceedings of the 2019 Conference on Empirical Methods in
  Natural Language Processing and the 9th International Joint Conference on
  Natural Language Processing (EMNLP-IJCNLP)}, pp.\  1962--1979, Hong Kong,
  China, November 2019{\natexlab{a}}. Association for Computational
  Linguistics.
\newblock \doi{10.18653/v1/D19-1204}.
\newblock URL \url{https://aclanthology.org/D19-1204}.

\bibitem[Yu et~al.(2019{\natexlab{b}})Yu, Zhang, Yasunaga, Tan, Lin, Li, Er,
  Li, Pang, Chen, Ji, Dixit, Proctor, Shim, Kraft, Zhang, Xiong, Socher, and
  Radev]{yu-etal-2019-sparc}
Tao Yu, Rui Zhang, Michihiro Yasunaga, Yi~Chern Tan, Xi~Victoria Lin, Suyi Li,
  Heyang Er, Irene Li, Bo~Pang, Tao Chen, Emily Ji, Shreya Dixit, David
  Proctor, Sungrok Shim, Jonathan Kraft, Vincent Zhang, Caiming Xiong, Richard
  Socher, and Dragomir Radev.
\newblock {SP}ar{C}: Cross-domain semantic parsing in context.
\newblock In \emph{Proceedings of the 57th Annual Meeting of the Association
  for Computational Linguistics}, pp.\  4511--4523, Florence, Italy, July
  2019{\natexlab{b}}. Association for Computational Linguistics.
\newblock \doi{10.18653/v1/P19-1443}.
\newblock URL \url{https://aclanthology.org/P19-1443}.

\end{thebibliography}
